\documentclass{article}

\pdfoutput=1
\PassOptionsToPackage{table}{xcolor}

\usepackage{microtype}
\usepackage{graphicx}
\usepackage{multirow}
\usepackage{booktabs} 

\usepackage{hyperref}

\usepackage[accepted]{icml2025}

\usepackage{amsmath}
\usepackage{amssymb}
\usepackage{mathtools}
\usepackage{amsthm}
\usepackage{subcaption}
\usepackage{wrapfig}

\definecolor{mygray}{gray}{0.6} 
\definecolor{lightblue}{RGB}{235, 245, 255}

\usepackage{colortbl}
\usepackage{enumitem}

\usepackage[capitalize,noabbrev]{cleveref}

\theoremstyle{plain}

\theoremstyle{definition}

\theoremstyle{remark}

\newcommand*{\method}{UniMoD}
\usepackage[textsize=tiny]{todonotes}

\icmltitlerunning{Submission and Formatting Instructions for ICML 2025}

\begin{document}

\twocolumn[
\icmltitle{UniMoD: Efficient Unified Multimodal Transformers with Mixture-of-Depths}



\icmlsetsymbol{equal}{*}

\begin{icmlauthorlist}
\icmlauthor{Weijia Mao}{sch,comp}
\icmlauthor{Zhenheng Yang}{comp}
\icmlauthor{Mike Zheng Shou}{sch}
\end{icmlauthorlist}

\icmlaffiliation{comp}{ByteDance}
\icmlaffiliation{sch}{{Show Lab, National University of Singapore}}

\icmlcorrespondingauthor{Mike Zheng Shou}{mike.zheng.shou@gmail.com}

\icmlkeywords{Machine Learning, ICML}

\vskip 0.3in
]


\printAffiliationsAndNotice{}  


\begin{abstract}

Unified multimodal transformers, which handle both generation and understanding tasks within a shared parameter space, have received increasing attention in recent research. Although various unified transformers have been proposed, training these models is costly due to redundant tokens and heavy attention computation. In the past, studies on large language models have demonstrated that token pruning methods, such as Mixture of Depths (MoD), can significantly improve computational efficiency. MoD employs a router to select the most important ones for processing within a transformer layer. However, directly applying MoD-based token pruning to unified transformers will result in suboptimal performance because different tasks exhibit varying levels of token redundancy. In our work, we analyze the unified transformers by (1) examining attention weight patterns, (2) evaluating the layer importance and token redundancy, and (3) analyzing task interactions. Our findings reveal that token redundancy is primarily influenced by different tasks and layers. Building on these findings, we introduce UniMoD, a \textbf{task-aware token pruning} method that employs a separate router for each task to determine which tokens should be pruned. We apply our method to Show-o and Emu3, reducing training FLOPs by approximately 15\% in Show-o and 40\% in Emu3, while maintaining or improving performance on several benchmarks. Code will be released at \url{https://github.com/showlab/UniMoD}.


\end{abstract}

\section{Introduction}
\label{sec:intro}

Unified multimodal transformers have received growing attention due to their ability to handle both generation and understanding tasks within a shared parameter space. Recent studies~\cite{showo,team2024chameleon,seed-x,emu3,sun2023emu,x-vila,janusflow,transfusion,luminamgpt} have explored different approaches for unified transformers. These models can be broadly categorized into two types: one where both generation and understanding tasks are handled using a fully autoregressive method~\cite{seed-x,team2024chameleon,emu3}, and another where generation tasks are addressed using diffusion or flow matching techniques, while autoregressive methods are used for understanding tasks~\cite{showo,transfusion,janusflow}. However, regardless of the approach, training these unified transformers remains time and memory intensive, primarily due to token redundancy and the computationally heavy attention mechanisms. In the past, few studies have explored efficient training strategies for these models. Therefore, developing efficient methods for training unified transformers remains a significant challenge.

\begin{figure*}
    \centering
    \includegraphics[width=\textwidth]{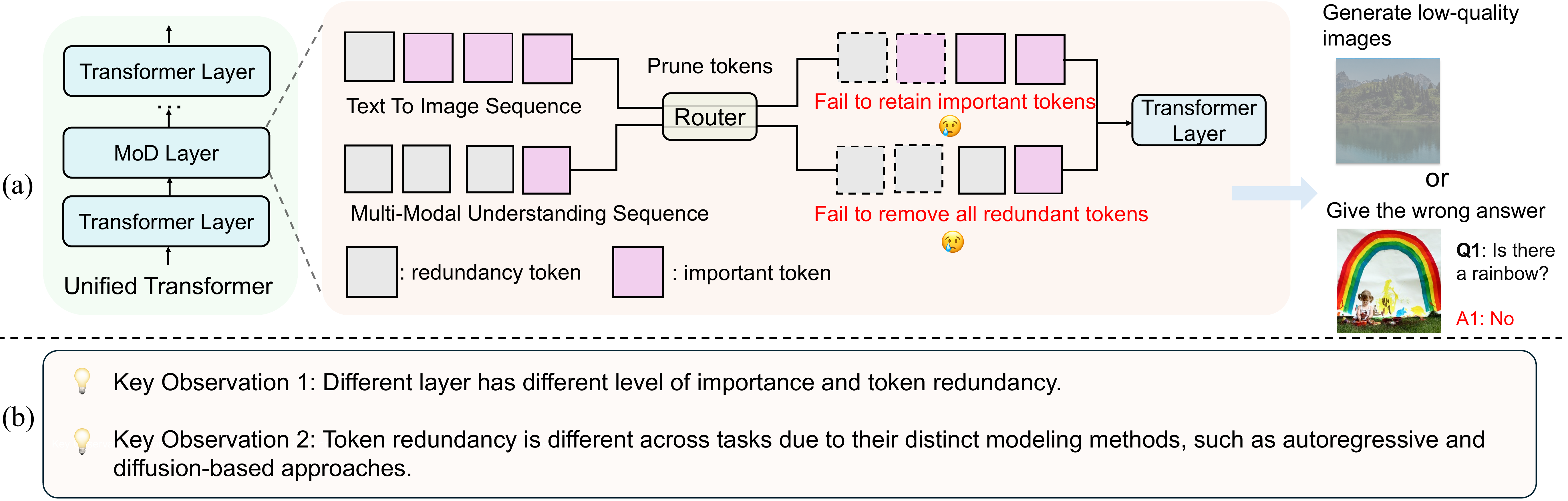}
    \vspace{-15pt} 
    \caption{
        (a) Pipeline and challenges of applying Mixture of Depths (MoD) to unified transformers. A single router prunes tokens across tasks and layers, leading to suboptimal performance due to inconsistent token redundancy.
        (b) Two key observations from our experiments on unified transformers, providing critical insights for our proposed method.
    }
    \label{fig:observation}
    \vspace{-12pt} 
\end{figure*}

The Mixture of Depths (MoD) approach has been employed in previous studies on large language models (LLMs)\cite{mod} and multimodal large language models (MLLMs)\cite{gamma-mod,videollmmod,pmod} to prune tokens. MoD employs a router to assign weights to tokens based on their significance, enabling the model to selectively prune less important tokens and reduce computational overhead. However, applying MoD to unified transformers presents unique challenges that are worth exploring. A straightforward application of MoD~\cite{moma} to unified transformers involves pruning tokens from sequences of any task, selecting and removing a fixed proportion of tokens at interleaved layer. 


However, as shown in Fig.~\ref{fig:observation}(a), this straightforward approach leads to suboptimal performance. A single router struggles to retain all important tokens and eliminate all redundant ones of all tasks. This limitation arises because different tasks exhibit varying levels of token redundancy at different layers. Consequently, employing a single router to uniformly prune tokens for both tasks is ineffective, as it cannot accommodate the distinct redundancy patterns inherent to each task. Therefore, without a thorough analysis of these factors, developing a token pruning strategy that meets the diverse requirements of unified transformer models remains challenging.

In this work, we conduct an empirical analysis of unified transformers from three different perspectives and present an effective solution. (1) We analyze attention weights in several unified transformers to explore whether different tasks and modalities influence attention weight patterns. (2) We evaluate layer importance and token redundancy to determine which layers should be pruned through a series of simple experiments. Firstly, our inference experiments demonstrate that different layers exhibit varying levels of importance. We then use the ARank metric~\cite{gamma-mod}, which refers to Attention Map Rank, to analyze the differences in token redundancy across layers and tasks. Higher values of ARank indicate lower token redundancy. (3) We explore the interactions between different tasks through two experiments. Firstly, we conduct experiments to assess how removing one task affects the benchmark performance of the other task. Additionally, we introduce a competitive setting where tokens from different tasks compete for selection, allowing us to compare their relative importance.


Based on our experimental results and analysis, we draw two important observations as shown in Fig.~\ref{fig:observation}(b). (1) Different layers within unified transformers exhibit varying levels of importance, with sequence importance and token redundancy fluctuating across layers.  (2) The token redundancy is different across different tasks due to distinct modeling methods. For example, in the Show-o model, generation tasks use diffusion-based approaches, while understanding tasks rely on autoregressive methods. These differences lead to varying levels of redundancy and importance in tokens of the same modality (e.g., image or text). Therefore, from these observations, we propose \method{}, a task-aware token pruning method for unified transformers. We transform several transformer layers into MoD blocks specialized for the generation task or the understanding task. Each task has its own router that assigns token weights, allowing the pruning process to adapt to each task. 


In our experiments, we select Show-o and Emu3 as representative unified transformers. Show-o handles generation and understanding tasks with distinct modeling approaches, while Emu3 employs a fully autoregressive approach for both tasks. Our method is general and applicable to various unified transformer architectures, regardless of how tasks are modeled. It reduces FLOPs by 15\% for Show-o and 40\% for Emu3, while maintaining or improving performance on certain benchmarks. Additionally, our approach extends to pure diffusion-based generation models like PixArt~\cite{pixart} and DiT~\cite{dit}, demonstrating the versatility and integration capability of our method across diverse generation frameworks.

Our main contributions are as follows.
\begin{itemize}[ itemsep=0em, topsep=0em]
    \itemsep0em 
    \item We conduct an empirical analysis of examining attention weights, layer importance and token redundancy, and the interactions between different tasks in the unified transformers.
    
    \item We identify that different modeling methods for different tasks lead to significant differences in token redundancy and importance across tasks. Furthermore, within the same task, different layers exhibit substantial variations in token redundancy and importance.
    
    \item To the best of our knowledge, we are the first work to propose a task-aware token pruning method for unified transformers, effectively reducing computation and memory usage while maintaining or improving performance.
    
\end{itemize}


    

\section{Related Work}
\label{sec:related_work}

\subsection{Unified Multi-modal Transformers}
Recently, there are several research work~\cite{seed-x,wu2023next,team2024chameleon,showo,lwm,sun2023emu,transfusion,CoDI,dreamllm,emu3,janusflow,luminamgpt,liquid,llamafusion,anil2023gemini,tokenflow,synergen,Orthus,omniflow} focusing on unified transformers that are capable of both generation and comprehension. Chameleon~\cite{team2024chameleon} and Emu3~\cite{emu3} employs an autoregressive approach for both generation and understanding tasks. SEED-X~\cite{seed-x} introduces a unified system for multimodal understanding and generation, incorporating a diffusion model alongside an LLM for generation tasks. Transfusion~\cite{transfusion} utilizes discrete tokens to represent texts and continuous embeddings to represent images. It integrates continuous diffusion methods with autoregressive approaches to handle generation and understanding tasks effectively. Show-o~\cite{showo} adopts discrete diffusion for generation and employs an autoregressive mode for understanding within a single model. JanusFlow~\cite{janusflow} combines the flow matching method for generation with the autoregressive method for understanding. However, all of these models require substantial resources for training.


\subsection{Sparse Computation for Transformers}
\textbf{Language Only.} Recently, Large Language Models (LLMs) have developed rapidly, resulting in the need for ever-increasing computational resources~\cite{llama,phi1.5,phi3,gpt3,gpt1}. As a result, much research~\cite{kvcache,layerskip,imageworth,attentionsink} focuses on sparse computation in LLMs. Mixture of Experts (MoE)\cite{surveymoe,openmoe,deepseekmoe,mixturalmoe} is a popular method that replaces the feed-forward (FFN) layers of transformer blocks with MoE layers, where input tokens are dynamically processed by the top-K experts via a router. However, MoE does not reduce training costs and is only efficient during inference. Mixture of Depths (MoD)\cite{mod} adopts a router at interleaved layer to decide whether tokens bypass the entire layer, thereby enhancing computational efficiency. Besides, there are several work using skip layers or early exit for sparse computation for LLMs~\cite{skiplayer,skipdecoder,earlyexit1}

\textbf{Multimodal Understanding and Generation.} 
MoE-LLaVA~\cite{moellava} and $\gamma$-MoD~\cite{gamma-mod} investigate Mixture of Experts (MoE) and Mixture of Depths (MoD) in multimodal large language models (MLLMs)\cite{llava,llava1.5,qwenvl,llava-next}, with $\gamma$-MoD introducing the ARank metric to assess token redundancy per layer. VideoLLM-MoD\cite{videollmmod} applies MoD to video LLMs~\cite{videollm}. MoMa~\cite{moma} integrates MoE and MoD into the Chameleon model~\cite{team2024chameleon}. However, it lacks results on generation tasks and most understanding benchmarks. Additionally, its application of MoD involves only a simplistic combination, without a design tailored for unified transformers. In the generation domain, concurrently, several studies~\cite{layermoddit,lazydit} have explored MoD methods for continous diffusion transformer models~\cite{dit}. In our work, we apply MoD to various unified transformers, demonstrating comparable or even improving results with fewer computational resources.

\begin{figure*}[htbp]
    \centering
    \includegraphics[width=\linewidth]{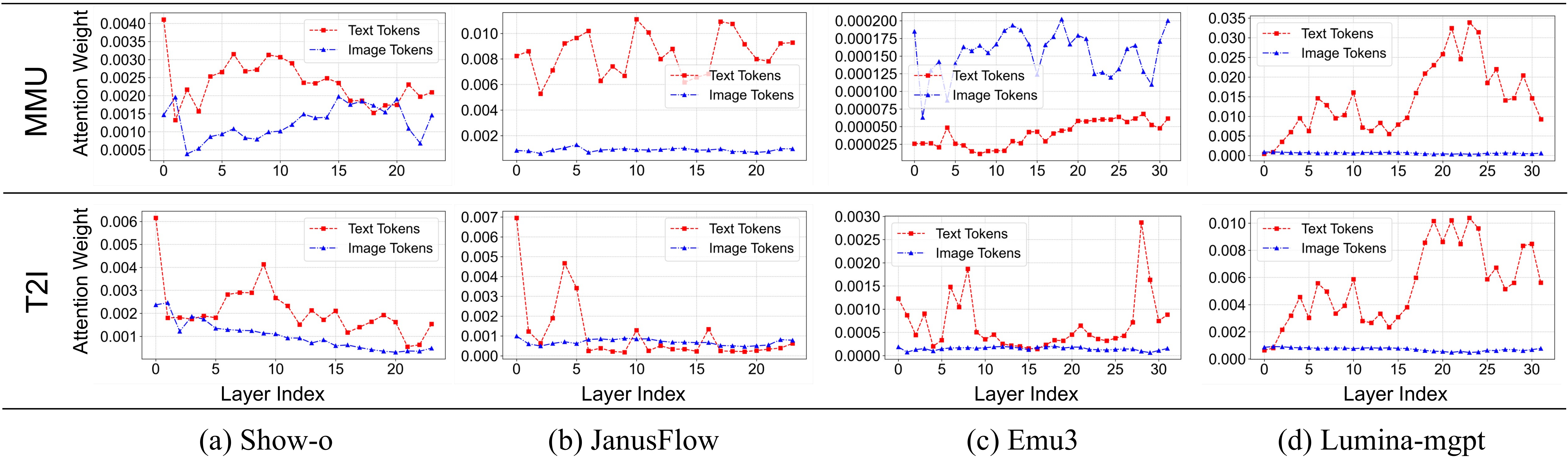}
    \vspace{-15pt}
    \caption{Attention weight for text and image tokens across different transformer layers for two tasks: Multi-Modal Understanding (MMU, top row) and Text-to-Image generation (T2I, bottom row). Each curve represents one token type, showing how attention allocation changes with model, layer index, and task. }
    
    \label{fig:attention_map_weights}
    \vspace{-10pt}
\end{figure*}

\section{Empirical Analysis of Unified Transformers}
In this section, we first introduce the unified transformers~\cite{showo,emu3} and the Mixture of Depths (MoD)\cite{mod} (Sec.\ref{sec:preliminary}). We then separately analyze the attention weights (Sec.\ref{sec:attention_map}), evaluate token redundancy in unified transformers (Sec.\ref{sec:topk_selection}), and analyze the interactions across different tasks (Sec.\ref{sec:SH_gumbel}). While Show-o serves as the main model for our experiments, we also investigate variations in other unified transformers~\cite{janusflow,luminamgpt,emu3} to gain a more comprehensive insight.

\subsection{Preliminary}
\label{sec:preliminary}


\textbf{Unified Multimodal Transformers}. Unified transformers typically come in two main types: one where both the generation and understanding tasks use AR-based methods, and the other where the generation task uses diffusion or flow matching while the understanding task employs autoregressive methods. We consider two representative unified multimodal transformer models: \textbf{Show-o}~\cite{showo} and \textbf{Emu3}~\cite{emu3}.

\textbf{Show-o} is a unified vision and language foundation model capable of both generation and comprehension. Based on the Phi language model~\cite{phi1.5}, Show-o adopts a discrete diffusion mode for generation and an autoregressive mode for understanding. To support both modeling approaches, Show-o employs two learning objectives:
Next Token Prediction (NTP): For generative tasks, given $M$ image tokens $\mathbf{u} = \{u_1, \ldots, u_M\}$ and $N$ text tokens $\mathbf{v} = \{v_1, \ldots, v_N\}$, the model maximizes the likelihood of the text tokens:
\begin{equation}
\mathcal{L}_{\text{NTP}} = \sum_{i=1}^{N} \log p_\theta(v_i \mid v_{1}, \ldots, v_{i-1}, u_{1}, \ldots, u_{M}).
\end{equation}
Mask Token Prediction (MTP): For comprehension tasks, given a sequence with masked image tokens $u_j^*$, the model maximizes the likelihood of the masked tokens:


\begin{equation}
\mathcal{L}_{\text{MTP}} = \sum_{j=1}^{M} \log p_\theta(u_j^* \mid \mathbf{u}_{\setminus j}, \mathbf{v}).
\end{equation}

\textbf{Emu3}, on the other hand, employs a fully autoregressive approach for both generation and multimodal understanding tasks. Its training objective is the standard next-token prediction across both modalities:

\begin{equation}
\mathcal{L}_{\text{AR}} = \sum_{i=1}^{N+M} \log p_\theta(x_i \mid x_{1}, \ldots, x_{i-1}),
\end{equation}
where $x_i$ represents the $i$-th token in the combined sequence of image and text tokens.

\textbf{Mixture of Depths (MoD)}~\cite{mod}. MoD is an efficient training method that reduces the number of tokens processed in each layer of a Transformer. Each layer employs a router that decides which tokens will be processed by that block. Tokens that are not chosen will skip the layer and proceed to the next layer. Thus, MoDs can be written as

\begin{equation}
x_i^{*} =
\begin{cases}
x_i + D(x_i) R(x_i), & \text{if } R(x_i) \geq \delta_s \\
x_i, & \text{if } R(x_i) < \delta_s
\end{cases}
.
\end{equation}

Formally, let $x_i \in \mathbb{R}^d$ denote the token vector in $x$, and let $\delta_s$ be the routing threshold at each layer. $D_i$ represents the $i$-th layer of the Transformer. $R_j(\cdot)$ is the corresponding routing function. Tokens that are not chosen will skip the layer, reducing computational cost. While Mixture of Experts (MoE)~\cite{deepseekmoe} is a scaling method that increases model capacity by activating only a subset of experts per token, it does not directly reduce training cost. On the other hand, MoD focuses on reducing the number of tokens processed at each layer by routing unnecessary tokens away, leading to a reduction in computational cost. Therefore, in our work we adopt MoD to train the unified transformers efficiently.


\subsection{Attention Weights}
\label{sec:attention_map}
We analyze the differences in attention weight patterns across tasks and modalities in unified transformers. We choose four unified models—Show-o~\cite{showo}, JanusFlow~\cite{janusflow}, Emu3~\cite{emu3} and Lumina-mgpt~\cite{luminamgpt}—to calculate the average attention weights received by image and text tokens across various layers. Based on the experimental results in Fig.~\ref{fig:attention_map_weights}, we draw an observation regarding the impact of different tasks on attention weight distribution.

\textbf{Observation 1}: \textit{Attention weight patterns of different modalities show significant differences depending on the task.}

As shown in Fig.~\ref{fig:attention_map_weights}, for Show-o, JanusFlow and Emu3, the attention weight patterns differ significantly between tasks. However, in Lumina-mgpt, the attention weight patterns are very similar across both tasks.

Different tasks lead to distinct modeling approaches and sequence organizations in unified transformers. Specifically, in the Show-o model, generation tasks utilize diffusion-based methods, while understanding tasks rely on autoregressive methods. Similarly, JanusFlow employs different modeling strategies for each task, resulting in varying attention weight patterns. Sequence organization refers to the relative positioning and arrangement of different modalities (e.g., text and image tokens) within the input sequence. In Emu3, although both tasks use autoregressive methods, differences in sequence organization lead to variations in modality performance. In contrast, Lumina-mgpt adopts an interleaved data training approach with consistent modeling and sequence design for both tasks, resulting in similar attention weight patterns across tasks.


Through observing the attention weight patterns, we realize that the importance of image and text tokens varies across tasks. Therefore, during pruning, we consider redundancy in tokens from all modalities, making the goal to prune tokens across both image and text modalities.





\begin{figure*}[t]
    \centering
    \includegraphics[width=1\linewidth]{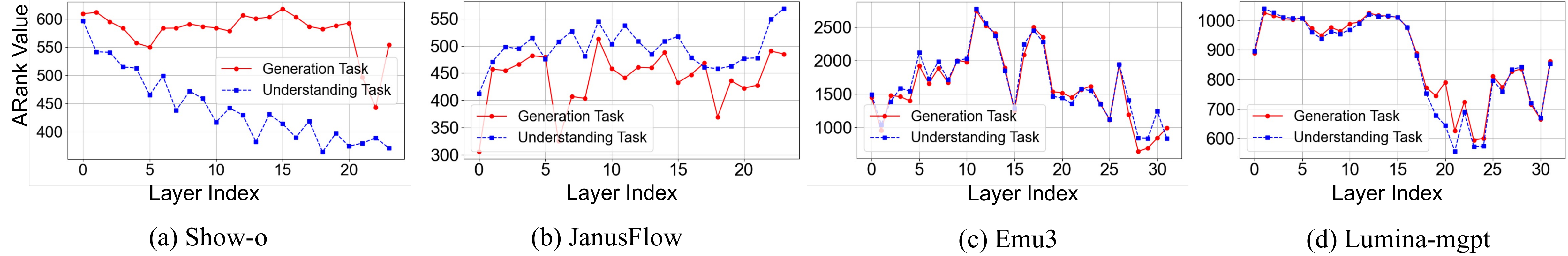}
    \vspace{-15pt}
    \caption{ARank variations across different layers for four unified transformers: Show-o, JanusFlow, Emu3, and Lumina-mgpt. ARank, defined as the rank of the attention map, represents sequence redundancy within each layer. Higher ARank values indicate lower sequence redundancy within each layer. }
    \label{fig:arank_showo}
    \vspace{-10pt}
\end{figure*}


\subsection{Layer Importance and Token Redundancy}
\label{sec:topk_selection}


We explore the significant variation in the importance of different layers and token redundancy for each task from two perspectives. First, we conduct simple inference experiments to analyze the benchmark performance in the Show-o model. Second, we use the ARank metric~\cite{gamma-mod} to evaluate token redundancy across layers in different unified transformers.


In the Show-o model, during the inference stage, we skip the odd-numbered layers and evaluate its performance on the GQA benchmark~\cite{gqa}, as shown in Tab.\ref{tab:inference_skip_layer} and we draw one observation.









\begin{table}
\centering
\caption{
GQA performance metrics observed when each specific layer is skipped during the inference process.
}
\vspace{-5pt}
\resizebox{1\linewidth}{!}{
\centering
\setlength{\tabcolsep}{1.8pt}

\begin{tabular}{lcccccccccccc}
\toprule
\multicolumn{1}{l}{layer}
&\multicolumn{1}{c}{ 1 }
&\multicolumn{1}{c}{ 3 }
&\multicolumn{1}{c}{ 5 }
&\multicolumn{1}{c}{ 7 }
&\multicolumn{1}{c}{ 9 }
&\multicolumn{1}{c}{ 11 }
&\multicolumn{1}{c}{ 13 }&\multicolumn{1}{c}{ 15 }&\multicolumn{1}{c}{ 17 }&\multicolumn{1}{c}{ 19 }&\multicolumn{1}{c}{ 21 }
&\multicolumn{1}{c}{ 23 }
  \\
   \midrule
  GQA
  &$35.0$
  &$0.0$
  &$48.0$
  &$49.0$ 
  &$49.4$
  &$50.0$
  &$50.2$
  &$51.7$
  &$51.5$
  &$51.4$ 
  &$51.1$
  &$50.9$
  \\ \bottomrule
\end{tabular}
}
\label{tab:inference_skip_layer}
\vspace{-10pt}
\end{table}

\textbf{Observation 2}: \textit{The contribution of each layer to the final outcome is different.}

As shown in Tab.~\ref{tab:inference_skip_layer}, GQA performance declines more significantly when tokens are skipped in the early layers compared to the late layers. This indicates that early layers are more critical for achieving optimal results.

Furthermore, we quantitatively assess the redundancy of tokens within each layer. Drawing inspiration from $\gamma$-mod~\cite{gamma-mod}, which utilizes the Attention Map Rank (ARank) metric to evaluate token-level redundancy in a layer, we apply this metric to analyze the unified transformers. ARank is calculated as the mean rank of the attention matrices:


\begin{equation}
\begin{aligned}
\tau(x_i, D_i) = \frac{1}{n_h} \sum_{h=1}^{n_h} \text{rank}(A_h),  
A_h = (W_{Qh} x_i)(W_{Kh} x_i)^\top.
\end{aligned}
\end{equation}

In these equations, \( \text{rank}(\cdot) \) denotes the matrix rank operation, \( n_h \) represents the number of attention heads, \( A_h \in \mathbb{R}^{l \times l} \) is the attention map for the \( h \)-th head, and \( W_{Qh} \in \mathbb{R}^{d \times d_h} \) and \( W_{Kh} \in \mathbb{R}^{d \times d_h} \) are the weight matrices for the query and key projections, respectively. This metric allows us to quantify the redundancy of tokens in each layer, providing insights into how different tasks and layers contribute to overall model efficiency. A layer with a low ARank means that most of its tokens are less informative. Based on the experimental results as shown in Fig.~\ref{fig:arank_showo}, we draw two observations regarding the token redundancy between different tasks.



\textbf{Observation 3}: \textit{The number of redundant tokens differs significantly between the generation and understanding tasks.}

As illustrated in Fig.~\ref{fig:arank_showo}, ARank values differ between tasks in the Show-o and JanusFlow models. In Show-o, the Text-to-Image(T2I) generation sequence has significantly higher ARank values than the Multi-Modal Understanding(MMU) sequence, indicating more redundant tokens in the MMU task. Conversely, Lumina-mgpt and Emu3 exhibit similar redundancy levels across both tasks. We attribute this difference to their modeling approaches: Show-o and JanusFlow use diffusion or flow matching for generation and autoregressive models for understanding, while Lumina-mgpt and Emu3 employ autoregressive methods for both tasks.

\textbf{Observation 4}: \textit{The redundancy of token sequences differs significantly across layers for different tasks.}

As shown in Figs.\ref{fig:arank_showo}, ARank values are significantly higher in the early layers compared to the later layers. In the Show-o model, the ARank value for the MMU task decreases as the layer index increases. In the T2I task, ARank values also show a substantial decrease in the later layers. All three models—JanusFlow, Lumina-mgpt, and Emu3—exhibit variations in ARank values across layers, indicating differing levels of token redundancy within each layer. We speculate these variations across layers may be related to the language models used.

Based on the analysis of layer importance and token redundancy, we conclude that pruning should focus on tokens in layers with high redundancy. The ARank metric can be used to identify such layers and determine the proportion of tokens to prune for both tasks.



\begin{figure*}
    \centering
    \includegraphics[width=1\linewidth]{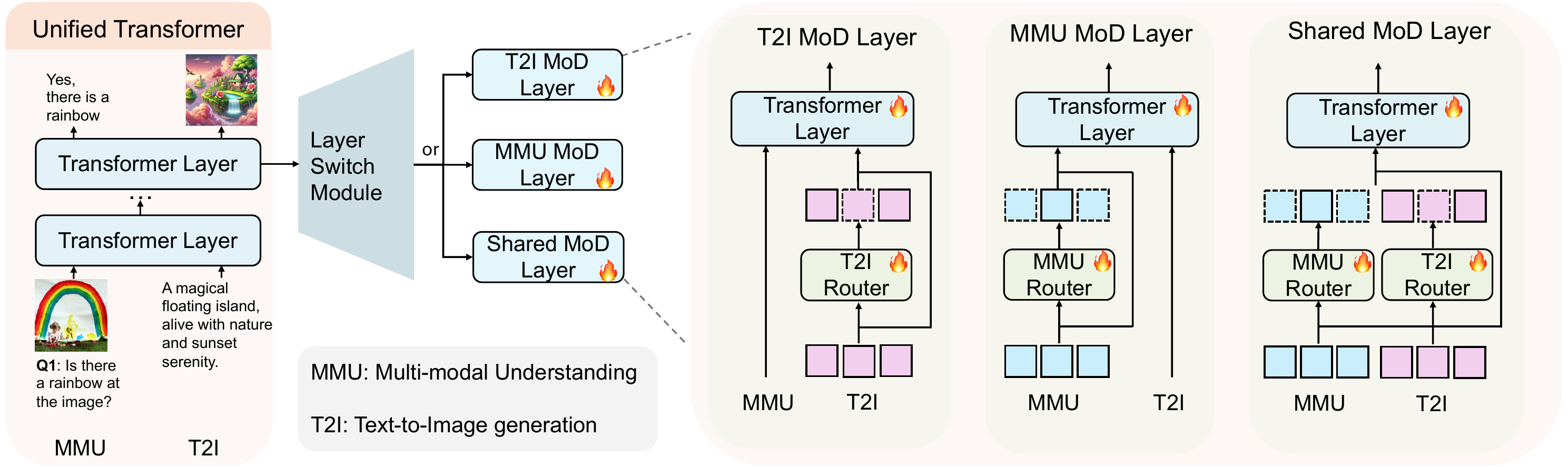}

    \caption{Pipeline of \method{}. The Layer Switch Module transforms dense transformer layers into three specialized types: T2I MoD layers for Text-to-Image (T2I) generation, MMU MoD layers for Multi-Modal Understanding (MMU), and Shared MoD layers for both tasks. For each task, task-aware routers with distinct capacities prune tokens of different modalities, thereby enhancing computational efficiency and maintaining performance across tasks.}

    \label{fig:pipeline}
    \vspace{-10pt}
\end{figure*}

\begin{figure}[h] 
    \centering
    \includegraphics[width=0.48\textwidth]{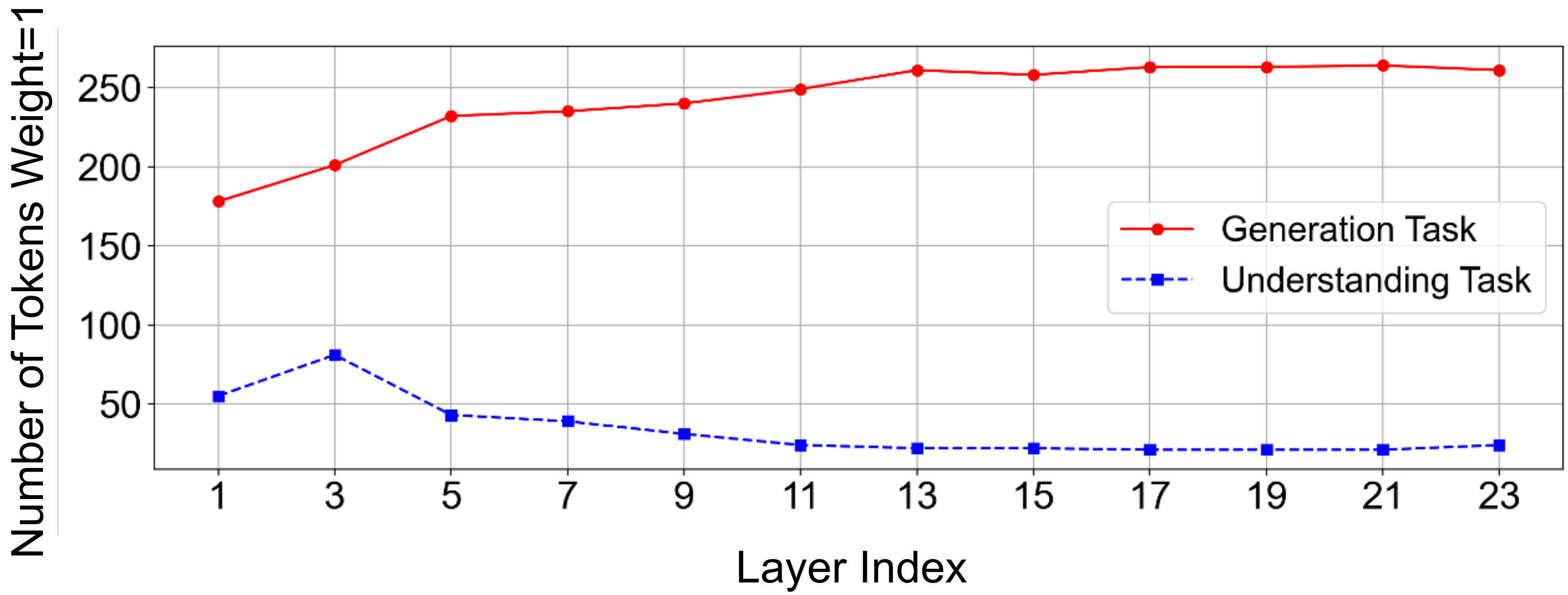}
    \vspace{-15pt}
    \caption{Token weight assignment using Gumbel Softmax. A higher number of tokens assigned a weight of 1 across layers indicates greater importance of generation task tokens compared to understanding task tokens.}
    \label{fig:gumbel_softmax}
    \vspace{-5pt}
\end{figure}


\subsection{Interactions Between Tasks}
\label{sec:SH_gumbel}


\textbf{Task Interactions in Unified Transformers.} We examine whether training multiple tasks simultaneously impacts performance using the Show-o model. Specifically, we assess the generation task and the understanding task under three configurations: training each task independently and training both concurrently. To evaluate performance, we utilize major understanding benchmarks—POPE~\cite{pope}, MME~\cite{mme}, GQA~\cite{gqa}, MMMU~\cite{mmmu}, VQAv2~\cite{vqav2}—and the GenEval~\cite{geneval} benchmark for generation tasks. As shown in Tab.~\ref{tab:comparison_showo}, the benchmark results for both tasks remained consistent across all training conditions. From these observations, we draw a key conclusion: \textbf{A large language model can effectively accommodate both tasks, as simultaneous training yields results comparable to training each task individually in the Show-o model.} The absence of a mutual enhancement effect may be attributed to Show-o modeling approach, which employs a discrete diffusion pipeline for generation and an autoregressive mode for understanding.













\begin{table}[t]
\caption{Benchmark results for three training configurations: `Show-o*' (simultaneously training Text-to-Image (T2I) and Multi-Modal Understanding (MMU) tasks), `Only T2I', and `Only MMU'.}
\vspace{-5pt}
\centering

\resizebox{1\linewidth}{!}{

\centering
\setlength{\tabcolsep}{1.8pt}

\begin{tabular}{l|ccccc|c}
\toprule

\multicolumn{1}{l|}{}
&\multicolumn{1}{c}{ MME$\uparrow$ }
&\multicolumn{1}{c}{ GQA$\uparrow$ }
&\multicolumn{1}{c}{ POPE$\uparrow$ }
&\multicolumn{1}{c}{ MMMU$\uparrow$ }
&\multicolumn{1}{c|}{ VQAv2$\uparrow$ }
&\multicolumn{1}{c}{ GenEval$\uparrow$ }

  \\
  \midrule
  Show-o*
  &$1032.0$
  &$52.5$
  &$77.9$ 
  &$27.4$
  &$62.4$ 
  &$0.63$

  \\Only MMU
  &$1030.0$
  &$52.2$
  &$77.7$ 
  &$26.7$
  &$61.7$ 
  &$-$

   \\Only T2I
  &$-$
  &$-$
  &$-$ 
  &$-$
  &$-$ 
  &$0.63$

\\ \bottomrule

\end{tabular}

}

\vspace{-10pt}
\label{tab:comparison_showo}
\end{table}


 
\textbf{Competitive Token Pruning Between Two Tasks.} We establish a competitive framework where tokens from Text-to-Image (T2I) and Multi-Modal Understanding (MMU) tasks vie for selection, allowing us to evaluate their relative importance. Utilizing the Straight-Through Gumbel-Softmax method, we assign binary weights (0 or 1) for pruning. To manage token redundancy, we set the router capacity to 0.5 and introduce an auxiliary loss to enforce budget constraints, ensuring each layer processes only half of the tokens from both tasks. The formula for the Straight-Through Gumbel-Softmax method is provided in the Appendix~\ref{gumbel_softmax_formula}. Based on Fig.~\ref{fig:gumbel_softmax}, we draw the following observation.


\textbf{Observation 5}: \textit{In the Show-o model, generation tokens are assigned higher weights, indicating that they contribute more significantly to the final results compared to understanding tokens.}

As shown in Fig.~\ref{fig:gumbel_softmax}, T2I sequence tokens are mostly assigned weights of 1, indicating they are largely retained. In contrast, MMU tokens receive lower weights, leading to higher pruning rates. This suggests that T2I tokens are more critical for loss optimization. We believe that uniformly processing tokens leads to an imbalance, causing one task to retain excessive tokens and potentially degrading the performance of the other. The interaction results in the Show-o model show mutual enhancement between tasks is minimal. These findings suggest pruning tokens separately for each task may result in more balanced and effective performance.







\section{Method}

\subsection{Task-Aware Mixture-of-Depths}

Based on our experimental results and analyses, we observe that token redundancy varies across tasks and layers. By analyzing the attention weights of unified transformers (see Sec.~\ref{sec:attention_map}), we find that the importance of image and text tokens varies across tasks. It suggests that token pruning should target tokens from all modalities. Our experiments with ARank values (Sec.\ref{sec:topk_selection}) across various layers and tasks show that redundancy levels depend on both the task and the specific layer. It suggests using the ARank metric to identify layers for pruning. Additionally, our task competition experiments (Sec.~\ref{sec:SH_gumbel}) reveal that token importance varies across tasks in terms of final loss optimization. It suggests tokens should be pruned separately for each task. Consequently, we introduce a task-aware token pruning method. The whole pipeline of \method{} is shown in Fig.~\ref{fig:pipeline}




\textbf{Task-Aware MoD Layer.} We transform dense transformer blocks into three specialized MoD blocks: the T2I MoD block for pruning tokens in the Text-to-Image (T2I) generation task, the MMU MoD block for pruning tokens in the Multi-Modal Understanding (MMU) task, and the Shared MoD block for pruning tokens in both tasks, as shown in Figure~\ref{fig:pipeline}. Specifically, the T2I MoD block only prunes T2I tokens while processing all MMU tokens, and the MMU MoD block only prunes MMU tokens while processing all T2I tokens. The Shared MoD block simultaneously prunes tokens from both tasks. For each task, we design dedicated routers with specific capacities, enabling the model to adaptively prune tokens based on task requirements. These routers target tokens from different modalities, ensuring efficient multi-modal processing and enhancing the model's overall computational efficiency.

\textbf{Layer Switch Module.} In this module, we utilize the ARank metric to identify which transformer layers should be transformed into MoD blocks. Specifically, during the training process, we calculate the ARank metric for each layer across different tasks using 50 samples per task. Based on the resulting ARank line chart, we select the top 12 layers with the highest ARank values for each respective task. The token pruning ratio is then determined proportionally based on the ARank values relative to the total sequence length. This approach allows us to dynamically prune tokens for each task, effectively managing token redundancy by leveraging the variance in ARank values and ensuring that pruning targets tokens from all modalities.

To accommodate different tasks, we design task-specific routers for each task (e.g., T2I and MMU) to perform different token pruning strategies. We also set task-specific thresholds. The formula is 

\begin{equation}
x_i^{*} =
\begin{cases}
x_i + D_t(x_i) R_t(x_i), & \text{if } R_t(x_i) \geq \delta_t \\
x_i, & \text{if } R_t(x_i) < \delta_t
\end{cases}
.
\end{equation}
\textbf{$x_i$} is the original token before pruning and \textbf{$x_i^{*}$} is the updated token after applying the pruning strategy. \textbf{$t$} represents the specific task (e.g., T2I, MMU). \textbf{$D_t(x_i)$} is the task-specific router function for task \( t \), determining how the token \( x_i \) should be adjusted. \textbf{$R_t(\cdot)$} is the corresponding routing function and \textbf{$R_t(x_i)$} is the task-specific weight for token \( x_i \) in task \( t \), indicating the token's relevance or significance. \textbf{$\delta_t$} is the task-specific threshold for task \( t \), deciding whether the token \( x_i \) should undergo pruning based on its importance score. This allows us to apply different token pruning criteria for each task, enabling dynamic adjustment of token pruning based on the specific requirements and ARank values of each task.



\section{Experiment}

\subsection{Implementation Details}

We choose Show-o~\cite{showo} and Emu-3~\cite{emu3} as representative models for our approach to token pruning, as they cover different types of unified transformers. Show-o integrates the diffusion pipeline with an autoregressive mode, addressing both generative and understanding tasks using distinct mechanisms. In contrast, Emu-3 employs the autoregressive mode for both tasks. By selecting these two models, we demonstrate that our method is applicable to a wide range of unified models.

In the Show-o model, when selecting layers to convert into MoD (Mixture of Depths) layers, we transform the last 12 layers into MoD layers for both tasks. For the Multi-Modal Understanding (MMU) task, we scale the capacity from 1 down to 0.2. For the Text-to-Image (T2I) task, we prune 20\% of the tokens in the later layers. We use a batch size of 10 for both the MMU and T2I tasks. For the T2I task, we use the image datasets from the original Show-o model, and for MMU, we employ the Cambrian dataset~\cite{cambrian}. Emu3 does not provide the training code or data required for MMU tasks. To validate our method, we employ the LLaVA-v1.5-mix-665K dataset to represent MMU capabilities and augment the training pipeline with additional MMU-specific code. For the T2I task, we use the same image data as the Show-o model. We finetune the model using 8 H100 GPUs, pruning 80\% of the tokens in each of the last 16 layers. Further details regarding the datasets and implementation are provided in the Appendix~\ref{showo_dataset}.

\subsection{Quantitative Results}

To evaluate multimodal understanding, we use the POPE\cite{pope}, MME~\cite{mme}, VQAv2~\cite{vqav2}, GQA~\cite{gqa}, and MMMU~\cite{mmmu} benchmarks. For generation capabilities, we rely on the GenEval~\cite{geneval} benchmark. To assess efficiency, we measure TFLOPs and training speed. Following the practice in DiT~\cite{dit}, we estimate the training compute as model TFLOPs × batch size × 3. The factor of 3 approximates the backward pass as requiring twice the compute of the forward pass.
















\begin{table}
\centering

\caption{Comparison of UniMoD with baseline methods across Multi-Modal Understanding (MMU) and Text-to-Image (T2I) benchmarks in the Show-o and Emu3 Models.}

\vspace{-5pt}
\resizebox{1\linewidth}{!}{

\centering
\setlength{\tabcolsep}{1.8pt}

\begin{tabular}{l|c|ccccc|c}
\toprule

\multicolumn{1}{l|}{}
&\multicolumn{1}{c|}{ TFLOPS$\downarrow$ }
&\multicolumn{1}{c}{ MME$\uparrow$ }
&\multicolumn{1}{c}{ GQA$\uparrow$ }
&\multicolumn{1}{c}{ POPE$\uparrow$ }
&\multicolumn{1}{c}{ MMMU$\uparrow$ }
&\multicolumn{1}{c|}{ VQAv2$\uparrow$ }
&\multicolumn{1}{c}{ GenEval$\uparrow$ }

  \\
  \midrule
  Show-o
  &$51.1$
  &$1056.0$
  &$56.3$
  &$79.8$ 
  &$25.8$
  &$68.3$ 
  &$0.62$

  \\LayerSkip
  &$25.6$
  &$828.4$
  &$45.6$
  &$74.2$ 
  &$27.2$
  &$53.9$ 
  &$0.29$

   \\EarlyExit
  &$25.6$
  &$947.0$
  &$51.1$
  &$79.5$ 
  &$24.1$
  &$60.4$ 
  &$0.26$

  \\
  \rowcolor{lightblue} 
  \method{}
  &$43.3$
  &$1093.7$
  &$54.5$
  &$80.3$ 
  &$25.7$
  &$66.2$ 
  &$0.61$
  \\
  \midrule
  Emu3
  &$89.0$
  &$881.3$
  &$46.0$
  &$76.0$ 
  &$25.1$
  &$54.8$ 
  &$0.46$

  \\
  \rowcolor{lightblue} 
  \method{}
  &$53.5$
  &$901.0$
  &$45.2$
  &$74.7$ 
  &$25.3$
  &$53.9$ 
  &$0.48$

\\ \bottomrule

\end{tabular}

}

\label{tab:table_baselines3}

\end{table}

\begin{table}[t]
\caption{Training Cost Comparison. It presents the computational requirements in terms of FLOPs and the training cost for both Text-to-Image (T2I) and Multi-Modal Understanding (MMU) tasks.}
\vspace{-5pt}
\centering
\resizebox{\linewidth}{!}{ 
\begin{tabular}{lcccccc}
    \toprule
    \multirow{3}{*}{Model} & \multirow{3}{*}{Params} & \multicolumn{2}{c}{T2I} & \multicolumn{2}{c}{MMU} \\
    \cmidrule(lr){3-4} \cmidrule(lr){5-6}
                           &                        & TFLOPs & Training Cost& TFLOPs & Training Cost \\
    \midrule
    Show-o      & 1.4B     & 51.1 &  1.30s/iter \& 67G &   51.1 & 1.30s/iter \& 67G  \\  
    \rowcolor{lightblue} 
    UniMod      & 1.4B       & 45.9 & 1.27s/iter \& 64G  & 40.8 & 1.25s/iter \& 61G  \\ 
    \midrule
    Emu3        & 8.5B       & 89.0  &  3.56s/iter \&  65G & 89.0  & 3.56s/iter \& 65G  \\ 
    \rowcolor{lightblue} 
    UniMod    & 8.5B      & 53.5 & 2.80s/iter \& 64G & 53.5 & 2.80s/iter \& 64G  \\ 
    \bottomrule
\end{tabular}
}

\vspace{-10pt}
\label{tab:training_costs}
\end{table}

\textbf{Baselines.} We compare our approach with various baselines, as shown in Tab.\ref{tab:table_baselines3}. \textit{Full Computation.} Using the vanilla training pipeline, all visual tokens and text tokens are passed through all transformer layers without any pruning. \textit{Early Exit.} Building on studies of language-only LLMs~\cite{earlyexit1}, we adopt a similar method for our unified models by performing an early exit at the 12th layer. \textit{LayerSkip.} Following the approach introduced in previous LLM studies~\cite{skiplayer}, we skip all tokens in the interleaved layers. This is equivalent to our method with a capacity of 1 in odd-numbered layers and a capacity of 0 in even-numbered layers. 

As shown in Tab.~\ref{tab:table_baselines3}, although LayerSkip and EarlyExit consume fewer TFLOPs, their performance is significantly inferior to the full computation model, especially for the T2I task. In contrast, our method achieves the best balance between performance and efficiency. We reduce the training FLOPs while maintaining comparable performance and even attaining better results on some benchmarks in the Show-o model. As shown in Tab.~\ref{tab:table_baselines3}, in the Emu model, our method reduces FLOPs by 40\%, while achieving comparable or better results compared to the full computation model in both T2I and MMU tasks. Our full computation results of Emu3 differ from the original paper because we use different training datasets due to the absence of relevant code and data in the public repository.


\textbf{Training Cost.} As shown in Tab.~\ref{tab:training_costs}, our method not only reduces FLOPs but also improves memory usage and training speed. In the Emu3 model, our approach is more efficient than Show-o in terms of FLOPs and training speed. We attribute this difference to the design of their image tokenizers: Emu3 uses 4096 tokens per image, while Show-o uses 1024 tokens. More tokens in Emu3 introduces more redundancy, thereby enhancing the efficiency of our method. The improvement in memory usage is less significant, likely due to Emu3's larger model size. Specific reasons are further discussed in the Appendix.~\ref{discussion_costs}.

\begin{table}
\caption{Ablation studies of \method{} in the Show-o model. `Basic MoD' directly integrates MoD into the unified transformer. `Interleaved layers' prunes tokens by selecting interleaved layers and `Single router' employs a single router to prune tokens for both tasks while selecting specific layers.}
\vspace{-5pt}
\centering

\resizebox{1\linewidth}{!}{

\centering
\setlength{\tabcolsep}{1.8pt}

\begin{tabular}{l|c|ccccc|c}

\toprule

\multicolumn{1}{l|}{}
&\multicolumn{1}{c|}{ TFLOPS$\downarrow$ }
&\multicolumn{1}{c}{ MME$\uparrow$ }
&\multicolumn{1}{c}{ GQA$\uparrow$ }
&\multicolumn{1}{c}{ POPE$\uparrow$ }
&\multicolumn{1}{c}{ MMMU$\uparrow$ }
&\multicolumn{1}{c|}{ VQAv2$\uparrow$ }
&\multicolumn{1}{c}{ GenEval$\uparrow$ }

  \\
  \midrule
  Basic MoD
  &$40.8$
  &$960.6$
  &$51.2$
  &$76.9$ 
  &$23.9$
  &$63.2$ 
  &$0.15$

  \\Interleaved layer
  &$43.3$
  &$920.3$
  &$52.1$
  &$74.7$ 
  &$25.1$
  &$63.2$ 
  &$0.50$

  \\Single router
  &$40.8$
  &$1052.0$
  &$54.4$
  &$80.2$ 
  &$25.6$
  &$65.5$ 
  &$0.50$

  \\
  \rowcolor{lightblue} 
  \method{}
  &$43.3$
  &$1093.7$
  &$54.5$
  &$80.3$ 
  &$25.7$
  &$66.2$ 
  &$0.61$

\\ \bottomrule

\end{tabular}

}



\vspace{-10pt}

\label{tab:table_ablation}
\end{table}

\subsection{Ablation Studies}



To evaluate the contribution of each design component in \method{}, we conduct an ablation study using the Show-o model, as shown in Tab.~\ref{tab:table_ablation}. We examine three configurations: (1) Basic MoD, which directly integrates MoD into the unified transformer; (2) Pruning Interleaved Layers, which selects interleaved layers for token pruning; and (3) Single Router, employing a single router to prune tokens for both tasks while selecting specific layers. Applying basic MoDs results in the poorest performance on both tasks. Using separate routers at interleaved layers slightly improves generation compared to basic MoDs, but outcomes remain suboptimal. Employing a single router for both tasks at specific layers slightly worsens understanding results and severely degrades generation performance. In contrast, our method achieves superior performance by effectively balancing efficiency and effectiveness. To ensure fairness, each ablation experiment maintains the same pruning rate as our method.

\subsection{Adaptation to Diffusion Models}

While our method is primarily designed for unified transformers, we also demonstrate its effectiveness in training and fine-tuning generation models such as DiT~\cite{dit} and PixArt~\cite{pixart}. Experimental results and implementation details are provided in the Appendix~\ref{adaption_to_diffusion}.

\section{Conclusion}



In this work, we present an efficient training method for unified transformers by analyzing attention weights, layer importance, and task interactions to identify sequence redundancy. Using these insights, we introduce a task-aware token pruning approach that reduces FLOPs while maintaining or enhancing performance across various benchmarks. Our method effectively balances efficiency and performance, demonstrating its applicability to unified transformers.

\section*{Impact Statement}
This paper presents work whose goal is to advance the field of Machine Learning. There are many potential societal
consequences of our work, none which we feel must be specifically highlighted here.

\nocite{langley00}

\bibliography{example_paper}

\begin{thebibliography}{144}
\providecommand{\natexlab}[1]{#1}
\providecommand{\url}[1]{\texttt{#1}}
\expandafter\ifx\csname urlstyle\endcsname\relax
  \providecommand{\doi}[1]{doi: #1}\else
  \providecommand{\doi}{doi: \begingroup \urlstyle{rm}\Url}\fi

\bibitem[Abdin et~al.(2024)Abdin, Jacobs, Awan, Aneja, Awadallah, Awadalla, Bach, Bahree, Bakhtiari, Behl, et~al.]{phi3}
Abdin, M., Jacobs, S.~A., Awan, A.~A., Aneja, J., Awadallah, A., Awadalla, H., Bach, N., Bahree, A., Bakhtiari, A., Behl, H., et~al.
\newblock Phi-3 technical report: A highly capable language model locally on your phone.
\newblock \emph{arXiv preprint arXiv:2404.14219}, 2024.

\bibitem[Aiello et~al.(2024)Aiello, YU, Nie, Aghajanyan, and Oguz]{jointly}
Aiello, E., YU, L., Nie, Y., Aghajanyan, A., and Oguz, B.
\newblock Jointly training large autoregressive multimodal models.
\newblock In \emph{ICLR}, 2024.

\bibitem[Alpher(2002)]{Alpher02}
Alpher, F.
\newblock Frobnication.
\newblock \emph{IEEE TPAMI}, 12\penalty0 (1):\penalty0 234--778, 2002.

\bibitem[Alpher \& Fotheringham-Smythe(2003)Alpher and Fotheringham-Smythe]{Alpher03}
Alpher, F. and Fotheringham-Smythe, F.
\newblock Frobnication revisited.
\newblock \emph{Journal of Foo}, 13\penalty0 (1):\penalty0 234--778, 2003.

\bibitem[Alpher \& Gamow(2005)Alpher and Gamow]{Alpher05}
Alpher, F. and Gamow, F.
\newblock Can a computer frobnicate?
\newblock In \emph{CVPR}, pp.\  234--778, 2005.

\bibitem[Alpher et~al.(2004)Alpher, Fotheringham-Smythe, and Gamow]{Alpher04}
Alpher, F., Fotheringham-Smythe, F., and Gamow, F.
\newblock Can a machine frobnicate?
\newblock \emph{Journal of Foo}, 14\penalty0 (1):\penalty0 234--778, 2004.

\bibitem[Anil et~al.(2023)Anil, Borgeaud, Wu, Alayrac, Yu, Soricut, Schalkwyk, Dai, Hauth, Millican, et~al.]{anil2023gemini}
Anil, R., Borgeaud, S., Wu, Y., Alayrac, J.-B., Yu, J., Soricut, R., Schalkwyk, J., Dai, A.~M., Hauth, A., Millican, K., et~al.
\newblock Gemini: A family of highly capable multimodal models.
\newblock \emph{arXiv preprint arXiv:2312.11805}, 1, 2023.

\bibitem[Austin et~al.(2021)Austin, Johnson, Ho, Tarlow, and Van Den~Berg]{d3pm}
Austin, J., Johnson, D.~D., Ho, J., Tarlow, D., and Van Den~Berg, R.
\newblock Structured denoising diffusion models in discrete state-spaces.
\newblock \emph{NeurIPS}, pp.\  17981--17993, 2021.

\bibitem[Bai et~al.(2023)Bai, Bai, Yang, Wang, Tan, Wang, Lin, Zhou, and Zhou]{qwenvl}
Bai, J., Bai, S., Yang, S., Wang, S., Tan, S., Wang, P., Lin, J., Zhou, C., and Zhou, J.
\newblock Qwen-vl: {A} frontier large vision-language model with versatile abilities.
\newblock \emph{CoRR}, abs/2308.12966, 2023.

\bibitem[Bao et~al.(2023)Bao, Nie, Xue, Cao, Li, Su, and Zhu]{uvit}
Bao, F., Nie, S., Xue, K., Cao, Y., Li, C., Su, H., and Zhu, J.
\newblock All are worth words: A vit backbone for diffusion models.
\newblock In \emph{CVPR}, 2023.

\bibitem[Bengio \& LeCun(2007)Bengio and LeCun]{Bengio+chapter2007}
Bengio, Y. and LeCun, Y.
\newblock Scaling learning algorithms towards {AI}.
\newblock In \emph{Large Scale Kernel Machines}. MIT Press, 2007.

\bibitem[Brown et~al.(2020)Brown, Mann, Ryder, Subbiah, Kaplan, Dhariwal, Neelakantan, Shyam, Sastry, Askell, Agarwal, Herbert{-}Voss, Krueger, Henighan, Child, Ramesh, Ziegler, Wu, Winter, Hesse, Chen, Sigler, Litwin, Gray, Chess, Clark, Berner, McCandlish, Radford, Sutskever, and Amodei]{gpt3}
Brown, T.~B., Mann, B., Ryder, N., Subbiah, M., Kaplan, J., Dhariwal, P., Neelakantan, A., Shyam, P., Sastry, G., Askell, A., Agarwal, S., Herbert{-}Voss, A., Krueger, G., Henighan, T., Child, R., Ramesh, A., Ziegler, D.~M., Wu, J., Winter, C., Hesse, C., Chen, M., Sigler, E., Litwin, M., Gray, S., Chess, B., Clark, J., Berner, C., McCandlish, S., Radford, A., Sutskever, I., and Amodei, D.
\newblock Language models are few-shot learners.
\newblock In \emph{NeurIPS}, 2020.

\bibitem[Cai et~al.(2024)Cai, Jiang, Wang, Tang, Kim, and Huang]{surveymoe}
Cai, W., Jiang, J., Wang, F., Tang, J., Kim, S., and Huang, J.
\newblock A survey on mixture of experts.
\newblock \emph{CoRR}, abs/2407.06204, 2024.

\bibitem[Campbell et~al.(2022)Campbell, Benton, De~Bortoli, Rainforth, Deligiannidis, and Doucet]{campbell2022continuous}
Campbell, A., Benton, J., De~Bortoli, V., Rainforth, T., Deligiannidis, G., and Doucet, A.
\newblock A continuous time framework for discrete denoising models.
\newblock \emph{NeurIPS}, pp.\  28266--28279, 2022.

\bibitem[Chang et~al.(2022)Chang, Zhang, Jiang, Liu, and Freeman]{chang2022maskgit}
Chang, H., Zhang, H., Jiang, L., Liu, C., and Freeman, W.~T.
\newblock Maskgit: Masked generative image transformer.
\newblock In \emph{CVPR}, pp.\  11315--11325, 2022.

\bibitem[Chang et~al.(2023)Chang, Zhang, Barber, Maschinot, Lezama, Jiang, Yang, Murphy, Freeman, Rubinstein, et~al.]{chang2023muse}
Chang, H., Zhang, H., Barber, J., Maschinot, A., Lezama, J., Jiang, L., Yang, M.-H., Murphy, K., Freeman, W.~T., Rubinstein, M., et~al.
\newblock Muse: Text-to-image generation via masked generative transformers.
\newblock \emph{arXiv preprint arXiv:2301.00704}, 2023.

\bibitem[Changpinyo et~al.(2021)Changpinyo, Sharma, Ding, and Soricut]{cc12m}
Changpinyo, S., Sharma, P., Ding, N., and Soricut, R.
\newblock Conceptual 12m: Pushing web-scale image-text pre-training to recognize long-tail visual concepts.
\newblock In \emph{CVPR}, pp.\  3558--3568, 2021.

\bibitem[Chen et~al.(2023{\natexlab{a}})Chen, Yu, Ge, Yao, Xie, Wu, Wang, Kwok, Luo, Lu, et~al.]{chen2023pixart}
Chen, J., Yu, J., Ge, C., Yao, L., Xie, E., Wu, Y., Wang, Z., Kwok, J., Luo, P., Lu, H., et~al.
\newblock Pixart-$\alpha$: Fast training of diffusion transformer for photorealistic text-to-image synthesis.
\newblock \emph{arXiv preprint arXiv:2310.00426}, 2023{\natexlab{a}}.

\bibitem[Chen et~al.(2024{\natexlab{a}})Chen, Lv, Wu, Lin, Song, Gao, Liu, Gao, Mao, and Shou]{videollm}
Chen, J., Lv, Z., Wu, S., Lin, K.~Q., Song, C., Gao, D., Liu, J., Gao, Z., Mao, D., and Shou, M.~Z.
\newblock Videollm-online: Online video large language model for streaming video.
\newblock In \emph{{CVPR}}, pp.\  18407--18418. {IEEE}, 2024{\natexlab{a}}.

\bibitem[Chen et~al.(2024{\natexlab{b}})Chen, Yu, Ge, Yao, Xie, Wang, Kwok, Luo, Lu, and Li]{pixart}
Chen, J., Yu, J., Ge, C., Yao, L., Xie, E., Wang, Z., Kwok, J.~T., Luo, P., Lu, H., and Li, Z.
\newblock Pixart-{\(\alpha\)}: Fast training of diffusion transformer for photorealistic text-to-image synthesis.
\newblock In \emph{{ICLR}}. OpenReview.net, 2024{\natexlab{b}}.

\bibitem[Chen et~al.(2023{\natexlab{b}})Chen, Li, Dong, Zhang, He, Wang, Zhao, and Lin]{sharegpt4v}
Chen, L., Li, J., Dong, X., Zhang, P., He, C., Wang, J., Zhao, F., and Lin, D.
\newblock Sharegpt4v: Improving large multi-modal models with better captions.
\newblock \emph{arXiv preprint arXiv:2311.12793}, 2023{\natexlab{b}}.

\bibitem[Chen et~al.(2024{\natexlab{c}})Chen, Zhao, Liu, Bai, Lin, Zhou, and Chang]{imageworth}
Chen, L., Zhao, H., Liu, T., Bai, S., Lin, J., Zhou, C., and Chang, B.
\newblock An image is worth 1/2 tokens after layer 2: Plug-and-play inference acceleration for large vision-language models.
\newblock \emph{CoRR}, abs/2403.06764, 2024{\natexlab{c}}.

\bibitem[Chen et~al.(2020)Chen, Radford, Child, Wu, Jun, Luan, and Sutskever]{chen2020generative}
Chen, M., Radford, A., Child, R., Wu, J., Jun, H., Luan, D., and Sutskever, I.
\newblock Generative pretraining from pixels.
\newblock In \emph{ICML}, pp.\  1691--1703, 2020.

\bibitem[Chowdhery et~al.(2023)Chowdhery, Narang, Devlin, Bosma, Mishra, Roberts, Barham, Chung, Sutton, Gehrmann, Schuh, Shi, Tsvyashchenko, Maynez, Rao, Barnes, Tay, Shazeer, Prabhakaran, Reif, Du, Hutchinson, Pope, Bradbury, Austin, Isard, Gur{-}Ari, Yin, Duke, Levskaya, Ghemawat, Dev, Michalewski, Garcia, Misra, Robinson, Fedus, Zhou, Ippolito, Luan, Lim, Zoph, Spiridonov, Sepassi, Dohan, Agrawal, Omernick, Dai, Pillai, Pellat, Lewkowycz, Moreira, Child, Polozov, Lee, Zhou, Wang, Saeta, Diaz, Firat, Catasta, Wei, Meier{-}Hellstern, Eck, Dean, Petrov, and Fiedel]{PALM}
Chowdhery, A., Narang, S., Devlin, J., Bosma, M., Mishra, G., Roberts, A., Barham, P., Chung, H.~W., Sutton, C., Gehrmann, S., Schuh, P., Shi, K., Tsvyashchenko, S., Maynez, J., Rao, A., Barnes, P., Tay, Y., Shazeer, N., Prabhakaran, V., Reif, E., Du, N., Hutchinson, B., Pope, R., Bradbury, J., Austin, J., Isard, M., Gur{-}Ari, G., Yin, P., Duke, T., Levskaya, A., Ghemawat, S., Dev, S., Michalewski, H., Garcia, X., Misra, V., Robinson, K., Fedus, L., Zhou, D., Ippolito, D., Luan, D., Lim, H., Zoph, B., Spiridonov, A., Sepassi, R., Dohan, D., Agrawal, S., Omernick, M., Dai, A.~M., Pillai, T.~S., Pellat, M., Lewkowycz, A., Moreira, E., Child, R., Polozov, O., Lee, K., Zhou, Z., Wang, X., Saeta, B., Diaz, M., Firat, O., Catasta, M., Wei, J., Meier{-}Hellstern, K., Eck, D., Dean, J., Petrov, S., and Fiedel, N.
\newblock Palm: Scaling language modeling with pathways.
\newblock \emph{J. Mach. Learn. Res.}, 24:\penalty0 240:1--240:113, 2023.

\bibitem[Dai et~al.(2024)Dai, Deng, Zhao, Xu, Gao, Chen, Li, Zeng, Yu, Wu, Xie, Li, Huang, Luo, Ruan, Sui, and Liang]{deepseekmoe}
Dai, D., Deng, C., Zhao, C., Xu, R.~X., Gao, H., Chen, D., Li, J., Zeng, W., Yu, X., Wu, Y., Xie, Z., Li, Y.~K., Huang, P., Luo, F., Ruan, C., Sui, Z., and Liang, W.
\newblock Deepseekmoe: Towards ultimate expert specialization in mixture-of-experts language models.
\newblock In \emph{{ACL} {(1)}}, pp.\  1280--1297. Association for Computational Linguistics, 2024.

\bibitem[Dai et~al.(2023)Dai, Li, Li, Tiong, Zhao, Wang, Li, Fung, and Hoi]{instructblip}
Dai, W., Li, J., Li, D., Tiong, A. M.~H., Zhao, J., Wang, W., Li, B., Fung, P., and Hoi, S.
\newblock Instructblip: Towards general-purpose vision-language models with instruction tuning, 2023.

\bibitem[Dehghani et~al.(2023)Dehghani, Djolonga, Mustafa, Padlewski, Heek, Gilmer, Steiner, Caron, Geirhos, Alabdulmohsin, Jenatton, Beyer, Tschannen, Arnab, Wang, Ruiz, Minderer, Puigcerver, Evci, Kumar, van Steenkiste, Elsayed, Mahendran, Yu, Oliver, Huot, Bastings, Collier, Gritsenko, Birodkar, Vasconcelos, Tay, Mensink, Kolesnikov, Pavetic, Tran, Kipf, Lucic, Zhai, Keysers, Harmsen, and Houlsby]{qknorm}
Dehghani, M., Djolonga, J., Mustafa, B., Padlewski, P., Heek, J., Gilmer, J., Steiner, A.~P., Caron, M., Geirhos, R., Alabdulmohsin, I., Jenatton, R., Beyer, L., Tschannen, M., Arnab, A., Wang, X., Ruiz, C.~R., Minderer, M., Puigcerver, J., Evci, U., Kumar, M., van Steenkiste, S., Elsayed, G.~F., Mahendran, A., Yu, F., Oliver, A., Huot, F., Bastings, J., Collier, M., Gritsenko, A.~A., Birodkar, V., Vasconcelos, C.~N., Tay, Y., Mensink, T., Kolesnikov, A., Pavetic, F., Tran, D., Kipf, T., Lucic, M., Zhai, X., Keysers, D., Harmsen, J.~J., and Houlsby, N.
\newblock Scaling vision transformers to 22 billion parameters.
\newblock In \emph{ICML}, pp.\  7480--7512, 2023.

\bibitem[Del~Corro et~al.(2023)Del~Corro, Del~Giorno, Agarwal, Yu, Awadallah, and Mukherjee]{skipdecoder}
Del~Corro, L., Del~Giorno, A., Agarwal, S., Yu, B., Awadallah, A., and Mukherjee, S.
\newblock Skipdecode: Autoregressive skip decoding with batching and caching for efficient llm inference.
\newblock \emph{arXiv preprint arXiv:2307.02628}, 2023.

\bibitem[Deng et~al.(2009)Deng, Dong, Socher, Li, Li, and Fei-Fei]{imagenet}
Deng, J., Dong, W., Socher, R., Li, L.-J., Li, K., and Fei-Fei, L.
\newblock Imagenet: A large-scale hierarchical image database.
\newblock In \emph{CVPR}, pp.\  248--255, 2009.

\bibitem[Dong et~al.(2023)Dong, Han, Peng, Qi, Ge, Yang, Zhao, Sun, Zhou, Wei, et~al.]{dong2023dreamllm}
Dong, R., Han, C., Peng, Y., Qi, Z., Ge, Z., Yang, J., Zhao, L., Sun, J., Zhou, H., Wei, H., et~al.
\newblock Dreamllm: Synergistic multimodal comprehension and creation.
\newblock \emph{arXiv preprint arXiv:2309.11499}, 2023.

\bibitem[Dong et~al.(2024)Dong, Han, Peng, Qi, Ge, Yang, Zhao, Sun, Zhou, Wei, Kong, Zhang, Ma, and Yi]{dreamllm}
Dong, R., Han, C., Peng, Y., Qi, Z., Ge, Z., Yang, J., Zhao, L., Sun, J., Zhou, H., Wei, H., Kong, X., Zhang, X., Ma, K., and Yi, L.
\newblock Dream{LLM}: Synergistic multimodal comprehension and creation.
\newblock In \emph{ICLR}, 2024.

\bibitem[Dosovitskiy et~al.(2021)Dosovitskiy, Beyer, Kolesnikov, Weissenborn, Zhai, Unterthiner, Dehghani, Minderer, Heigold, Gelly, Uszkoreit, and Houlsby]{vit}
Dosovitskiy, A., Beyer, L., Kolesnikov, A., Weissenborn, D., Zhai, X., Unterthiner, T., Dehghani, M., Minderer, M., Heigold, G., Gelly, S., Uszkoreit, J., and Houlsby, N.
\newblock An image is worth 16x16 words: Transformers for image recognition at scale.
\newblock In \emph{{ICLR}}. OpenReview.net, 2021.

\bibitem[Elhoushi et~al.(2024{\natexlab{a}})Elhoushi, Shrivastava, Liskovich, Hosmer, Wasti, Lai, Mahmoud, Acun, Agarwal, Roman, et~al.]{layerskip}
Elhoushi, M., Shrivastava, A., Liskovich, D., Hosmer, B., Wasti, B., Lai, L., Mahmoud, A., Acun, B., Agarwal, S., Roman, A., et~al.
\newblock Layer skip: Enabling early exit inference and self-speculative decoding.
\newblock \emph{arXiv preprint arXiv:2404.16710}, 2024{\natexlab{a}}.

\bibitem[Elhoushi et~al.(2024{\natexlab{b}})Elhoushi, Shrivastava, Liskovich, Hosmer, Wasti, Lai, Mahmoud, Acun, Agarwal, Roman, et~al.]{skiplayer}
Elhoushi, M., Shrivastava, A., Liskovich, D., Hosmer, B., Wasti, B., Lai, L., Mahmoud, A., Acun, B., Agarwal, S., Roman, A., et~al.
\newblock Layer skip: Enabling early exit inference and self-speculative decoding.
\newblock \emph{arXiv preprint arXiv:2404.16710}, 2024{\natexlab{b}}.

\bibitem[Esser et~al.(2021{\natexlab{a}})Esser, Rombach, and Ommer]{taming}
Esser, P., Rombach, R., and Ommer, B.
\newblock Taming transformers for high-resolution image synthesis.
\newblock In \emph{CVPR}, pp.\  12873--12883, 2021{\natexlab{a}}.

\bibitem[Esser et~al.(2021{\natexlab{b}})Esser, Rombach, and Ommer]{vqgan}
Esser, P., Rombach, R., and Ommer, B.
\newblock Taming transformers for high-resolution image synthesis.
\newblock In \emph{CVPR}, pp.\  12873--12883, 2021{\natexlab{b}}.

\bibitem[Esser et~al.(2024)Esser, Kulal, Blattmann, Entezari, M{\"u}ller, Saini, Levi, Lorenz, Sauer, Boesel, et~al.]{sd3}
Esser, P., Kulal, S., Blattmann, A., Entezari, R., M{\"u}ller, J., Saini, H., Levi, Y., Lorenz, D., Sauer, A., Boesel, F., et~al.
\newblock Scaling rectified flow transformers for high-resolution image synthesis.
\newblock In \emph{ICML}, 2024.

\bibitem[Fu et~al.(2023)Fu, Chen, Shen, Qin, Zhang, Lin, Qiu, Lin, Yang, Zheng, Li, Sun, and Ji]{mme}
Fu, C., Chen, P., Shen, Y., Qin, Y., Zhang, M., Lin, X., Qiu, Z., Lin, W., Yang, J., Zheng, X., Li, K., Sun, X., and Ji, R.
\newblock {MME:} {A} comprehensive evaluation benchmark for multimodal large language models.
\newblock \emph{CoRR}, abs/2306.13394, 2023.

\bibitem[Ge et~al.(2024{\natexlab{a}})Ge, Zhang, Liu, Zhang, Han, and Gao]{kvcache}
Ge, S., Zhang, Y., Liu, L., Zhang, M., Han, J., and Gao, J.
\newblock Model tells you what to discard: Adaptive {KV} cache compression for llms.
\newblock In \emph{{ICLR}}. OpenReview.net, 2024{\natexlab{a}}.

\bibitem[Ge et~al.(2024{\natexlab{b}})Ge, Zhao, Zhu, Ge, Yi, Song, Li, Ding, and Shan]{seed-x}
Ge, Y., Zhao, S., Zhu, J., Ge, Y., Yi, K., Song, L., Li, C., Ding, X., and Shan, Y.
\newblock Seed-x: Multimodal models with unified multi-granularity comprehension and generation.
\newblock \emph{arXiv preprint arXiv:2404.14396}, 2024{\natexlab{b}}.

\bibitem[Geva et~al.(2022)Geva, Caciularu, Wang, and Goldberg]{earlyexit1}
Geva, M., Caciularu, A., Wang, K.~R., and Goldberg, Y.
\newblock Transformer feed-forward layers build predictions by promoting concepts in the vocabulary space.
\newblock In \emph{{EMNLP}}, pp.\  30--45. Association for Computational Linguistics, 2022.

\bibitem[Ghazvininejad et~al.(2019)Ghazvininejad, Levy, Liu, and Zettlemoyer]{maskpredict}
Ghazvininejad, M., Levy, O., Liu, Y., and Zettlemoyer, L.
\newblock Mask-predict: Parallel decoding of conditional masked language models.
\newblock In \emph{EMNLP}, pp.\  6111--6120, 2019.

\bibitem[Ghosh et~al.(2023)Ghosh, Hajishirzi, and Schmidt]{geneval}
Ghosh, D., Hajishirzi, H., and Schmidt, L.
\newblock Geneval: An object-focused framework for evaluating text-to-image alignment.
\newblock In \emph{NeurIPS}, 2023.

\bibitem[Goodfellow et~al.(2014)Goodfellow, Pouget-Abadie, Mirza, Xu, Warde-Farley, Ozair, Courville, and Bengio]{gan}
Goodfellow, I., Pouget-Abadie, J., Mirza, M., Xu, B., Warde-Farley, D., Ozair, S., Courville, A., and Bengio, Y.
\newblock Generative adversarial nets.
\newblock \emph{NeurIPS}, 2014.

\bibitem[Goodfellow et~al.(2016)Goodfellow, Bengio, Courville, and Bengio]{goodfellow2016deep}
Goodfellow, I., Bengio, Y., Courville, A., and Bengio, Y.
\newblock \emph{Deep learning}, volume~1.
\newblock MIT Press, 2016.

\bibitem[Goyal et~al.(2017)Goyal, Khot, Summers{-}Stay, Batra, and Parikh]{vqav2}
Goyal, Y., Khot, T., Summers{-}Stay, D., Batra, D., and Parikh, D.
\newblock Making the {V} in {VQA} matter: Elevating the role of image understanding in visual question answering.
\newblock In \emph{{CVPR}}, pp.\  6325--6334. {IEEE} Computer Society, 2017.

\bibitem[Gu et~al.(2022)Gu, Chen, Bao, Wen, Zhang, Chen, Yuan, and Guo]{vqdiffusion}
Gu, S., Chen, D., Bao, J., Wen, F., Zhang, B., Chen, D., Yuan, L., and Guo, B.
\newblock Vector quantized diffusion model for text-to-image synthesis.
\newblock In \emph{CVPR}, pp.\  10696--10706, 2022.

\bibitem[Gu et~al.(2024)Gu, Wang, Ge, Shan, and Shou]{gu2024rethinking}
Gu, Y., Wang, X., Ge, Y., Shan, Y., and Shou, M.~Z.
\newblock Rethinking the objectives of vector-quantized tokenizers for image synthesis.
\newblock In \emph{CVPR}, pp.\  7631--7640, 2024.

\bibitem[Hinton et~al.(2006)Hinton, Osindero, and Teh]{Hinton06}
Hinton, G.~E., Osindero, S., and Teh, Y.~W.
\newblock A fast learning algorithm for deep belief nets.
\newblock \emph{Neural Computation}, 18:\penalty0 1527--1554, 2006.

\bibitem[Ho \& Salimans(2022)Ho and Salimans]{ho2022classifier}
Ho, J. and Salimans, T.
\newblock Classifier-free diffusion guidance.
\newblock \emph{arXiv preprint arXiv:2207.12598}, 2022.

\bibitem[Ho et~al.(2020{\natexlab{a}})Ho, Jain, and Abbeel]{ddpm}
Ho, J., Jain, A., and Abbeel, P.
\newblock Denoising diffusion probabilistic models.
\newblock In \emph{NeurIPS}, pp.\  6840--6851, 2020{\natexlab{a}}.

\bibitem[Ho et~al.(2020{\natexlab{b}})Ho, Jain, and Abbeel]{ho2020denoising}
Ho, J., Jain, A., and Abbeel, P.
\newblock Denoising diffusion probabilistic models.
\newblock \emph{NeurIPS}, pp.\  6840--6851, 2020{\natexlab{b}}.

\bibitem[Ho et~al.(2022)Ho, Salimans, Gritsenko, Chan, Norouzi, and Fleet]{ho2022video}
Ho, J., Salimans, T., Gritsenko, A., Chan, W., Norouzi, M., and Fleet, D.~J.
\newblock Video diffusion models.
\newblock \emph{NeurIPS}, 2022.

\bibitem[Hoogeboom et~al.(2022)Hoogeboom, Gritsenko, Bastings, Poole, van~den Berg, and Salimans]{ARDM}
Hoogeboom, E., Gritsenko, A.~A., Bastings, J., Poole, B., van~den Berg, R., and Salimans, T.
\newblock Autoregressive diffusion models.
\newblock In \emph{{ICLR}}. OpenReview.net, 2022.

\bibitem[Hu et~al.(2023)Hu, Zheng, Yang, Cham, Zheng, Wang, Tao, and Suganthan]{hu2023unified}
Hu, M., Zheng, C., Yang, Z., Cham, T.-J., Zheng, H., Wang, C., Tao, D., and Suganthan, P.~N.
\newblock Unified discrete diffusion for simultaneous vision-language generation.
\newblock In \emph{ICLR}, 2023.

\bibitem[Hudson \& Manning(2019)Hudson and Manning]{gqa}
Hudson, D.~A. and Manning, C.~D.
\newblock {GQA:} {A} new dataset for real-world visual reasoning and compositional question answering.
\newblock In \emph{{CVPR}}, pp.\  6700--6709. Computer Vision Foundation / {IEEE}, 2019.

\bibitem[Jiang et~al.(2024)Jiang, Sablayrolles, Roux, Mensch, Savary, Bamford, Chaplot, de~Las~Casas, Hanna, Bressand, Lengyel, Bour, Lample, Lavaud, Saulnier, Lachaux, Stock, Subramanian, Yang, Antoniak, Scao, Gervet, Lavril, Wang, Lacroix, and Sayed]{mixturalmoe}
Jiang, A.~Q., Sablayrolles, A., Roux, A., Mensch, A., Savary, B., Bamford, C., Chaplot, D.~S., de~Las~Casas, D., Hanna, E.~B., Bressand, F., Lengyel, G., Bour, G., Lample, G., Lavaud, L.~R., Saulnier, L., Lachaux, M., Stock, P., Subramanian, S., Yang, S., Antoniak, S., Scao, T.~L., Gervet, T., Lavril, T., Wang, T., Lacroix, T., and Sayed, W.~E.
\newblock Mixtral of experts.
\newblock \emph{CoRR}, abs/2401.04088, 2024.

\bibitem[Kang et~al.(2023)Kang, Zhu, Zhang, Park, Shechtman, Paris, and Park]{gigagan}
Kang, M., Zhu, J., Zhang, R., Park, J., Shechtman, E., Paris, S., and Park, T.
\newblock Scaling up gans for text-to-image synthesis.
\newblock In \emph{{CVPR}}, pp.\  10124--10134. {IEEE}, 2023.

\bibitem[Kingma \& Welling(2013)Kingma and Welling]{vae}
Kingma, D.~P. and Welling, M.
\newblock Auto-encoding variational bayes.
\newblock \emph{arXiv preprint arXiv:1312.6114}, 2013.

\bibitem[Kirillov et~al.(2023)Kirillov, Mintun, Ravi, Mao, Rolland, Gustafson, Xiao, Whitehead, Berg, Lo, et~al.]{sa1b}
Kirillov, A., Mintun, E., Ravi, N., Mao, H., Rolland, C., Gustafson, L., Xiao, T., Whitehead, S., Berg, A.~C., Lo, W.-Y., et~al.
\newblock Segment anything.
\newblock In \emph{ICCV}, pp.\  4015--4026, 2023.

\bibitem[Kondratyuk et~al.(2023)Kondratyuk, Yu, Gu, Lezama, Huang, Hornung, Adam, Akbari, Alon, Birodkar, et~al.]{kondratyuk2023videopoet}
Kondratyuk, D., Yu, L., Gu, X., Lezama, J., Huang, J., Hornung, R., Adam, H., Akbari, H., Alon, Y., Birodkar, V., et~al.
\newblock Videopoet: A large language model for zero-shot video generation.
\newblock \emph{arXiv preprint arXiv:2312.14125}, 2023.

\bibitem[Kou et~al.(2024)Kou, Jin, Liu, Ma, Jia, Chen, Jiang, and Deng]{Orthus}
Kou, S., Jin, J., Liu, C., Ma, Y., Jia, J., Chen, Q., Jiang, P., and Deng, Z.
\newblock Orthus: Autoregressive interleaved image-text generation with modality-specific heads.
\newblock \emph{CoRR}, abs/2412.00127, 2024.

\bibitem[LastName(2014{\natexlab{a}})]{Authors14}
LastName, F.
\newblock The frobnicatable foo filter, 2014{\natexlab{a}}.
\newblock Face and Gesture submission ID 324. Supplied as supplemental material {\tt fg324.pdf}.

\bibitem[LastName(2014{\natexlab{b}})]{Authors14b}
LastName, F.
\newblock Frobnication tutorial, 2014{\natexlab{b}}.
\newblock Supplied as supplemental material {\tt tr.pdf}.

\bibitem[Li et~al.(2024{\natexlab{a}})Li, Gan, Yang, Yang, Li, Wang, Gao, et~al.]{li2024multimodal}
Li, C., Gan, Z., Yang, Z., Yang, J., Li, L., Wang, L., Gao, J., et~al.
\newblock Multimodal foundation models: From specialists to general-purpose assistants.
\newblock \emph{Foundations and Trends{\textregistered} in Computer Graphics and Vision}, 16\penalty0 (1-2):\penalty0 1--214, 2024{\natexlab{a}}.

\bibitem[Li et~al.(2024{\natexlab{b}})Li, Tian, Shao, Zhu, Wang, Zhu, Dou, Wang, Li, Lu, and Dai]{synergen}
Li, H., Tian, C., Shao, J., Zhu, X., Wang, Z., Zhu, J., Dou, W., Wang, X., Li, H., Lu, L., and Dai, J.
\newblock Synergen-vl: Towards synergistic image understanding and generation with vision experts and token folding.
\newblock \emph{CoRR}, abs/2412.09604, 2024{\natexlab{b}}.

\bibitem[Li et~al.(2023{\natexlab{a}})Li, Li, Savarese, and Hoi]{blip2}
Li, J., Li, D., Savarese, S., and Hoi, S. C.~H.
\newblock {BLIP-2:} bootstrapping language-image pre-training with frozen image encoders and large language models.
\newblock In \emph{{ICML}}, pp.\  19730--19742, 2023{\natexlab{a}}.

\bibitem[Li et~al.(2024{\natexlab{c}})Li, Kallidromitis, Gokul, Liao, Kato, Kozuka, and Grover]{omniflow}
Li, S., Kallidromitis, K., Gokul, A., Liao, Z., Kato, Y., Kozuka, K., and Grover, A.
\newblock Omniflow: Any-to-any generation with multi-modal rectified flows.
\newblock \emph{CoRR}, abs/2412.01169, 2024{\natexlab{c}}.

\bibitem[Li et~al.(2023{\natexlab{b}})Li, Bubeck, Eldan, Del~Giorno, Gunasekar, and Lee]{phi1.5}
Li, Y., Bubeck, S., Eldan, R., Del~Giorno, A., Gunasekar, S., and Lee, Y.~T.
\newblock Textbooks are all you need ii: phi-1.5 technical report.
\newblock \emph{arXiv preprint arXiv:2309.05463}, 2023{\natexlab{b}}.

\bibitem[Li et~al.(2023{\natexlab{c}})Li, Du, Zhou, Wang, Zhao, and Wen]{pope}
Li, Y., Du, Y., Zhou, K., Wang, J., Zhao, W.~X., and Wen, J.
\newblock Evaluating object hallucination in large vision-language models.
\newblock In \emph{{EMNLP}}, pp.\  292--305. Association for Computational Linguistics, 2023{\natexlab{c}}.

\bibitem[Lin et~al.(2024{\natexlab{a}})Lin, Tang, Ye, Cui, Zhu, Jin, Zhang, Ning, and Yuan]{moellava}
Lin, B., Tang, Z., Ye, Y., Cui, J., Zhu, B., Jin, P., Zhang, J., Ning, M., and Yuan, L.
\newblock Moe-llava: Mixture of experts for large vision-language models.
\newblock \emph{CoRR}, abs/2401.15947, 2024{\natexlab{a}}.

\bibitem[Lin et~al.(2024{\natexlab{b}})Lin, Shrivastava, Luo, Iyer, Lewis, Ghosh, Zettlemoyer, and Aghajanyan]{moma}
Lin, X.~V., Shrivastava, A., Luo, L., Iyer, S., Lewis, M., Ghosh, G., Zettlemoyer, L., and Aghajanyan, A.
\newblock Moma: Efficient early-fusion pre-training with mixture of modality-aware experts.
\newblock \emph{CoRR}, abs/2407.21770, 2024{\natexlab{b}}.

\bibitem[Liu et~al.(2024{\natexlab{a}})Liu, Zhao, Zhuo, Lin, Qiao, Li, and Gao]{luminamgpt}
Liu, D., Zhao, S., Zhuo, L., Lin, W., Qiao, Y., Li, H., and Gao, P.
\newblock Lumina-mgpt: Illuminate flexible photorealistic text-to-image generation with multimodal generative pretraining.
\newblock \emph{CoRR}, abs/2408.02657, 2024{\natexlab{a}}.

\bibitem[Liu et~al.(2024{\natexlab{b}})Liu, Li, Li, and Lee]{llava1.5}
Liu, H., Li, C., Li, Y., and Lee, Y.~J.
\newblock Improved baselines with visual instruction tuning.
\newblock In \emph{CVPR}, pp.\  26296--26306, 2024{\natexlab{b}}.

\bibitem[Liu et~al.(2024{\natexlab{c}})Liu, Li, Li, Li, Zhang, Shen, and Lee]{llava-next}
Liu, H., Li, C., Li, Y., Li, B., Zhang, Y., Shen, S., and Lee, Y.~J.
\newblock Llava-next: Improved reasoning, ocr, and world knowledge, 2024{\natexlab{c}}.

\bibitem[Liu et~al.(2024{\natexlab{d}})Liu, Li, Wu, and Lee]{llava}
Liu, H., Li, C., Wu, Q., and Lee, Y.~J.
\newblock Visual instruction tuning.
\newblock \emph{NeurIPS}, 36, 2024{\natexlab{d}}.

\bibitem[Liu et~al.(2024{\natexlab{e}})Liu, Yan, Zaharia, and Abbeel]{lwm}
Liu, H., Yan, W., Zaharia, M., and Abbeel, P.
\newblock World model on million-length video and language with ringattention.
\newblock \emph{arXiv preprint}, 2024{\natexlab{e}}.

\bibitem[Luo et~al.(2024)Luo, Luo, Ji, Zhou, Sun, Shen, and Ji]{gamma-mod}
Luo, Y., Luo, G., Ji, J., Zhou, Y., Sun, X., Shen, Z., and Ji, R.
\newblock gamma-mod: Exploring mixture-of-depth adaptation for multimodal large language models.
\newblock \emph{arXiv preprint arXiv:2410.13859}, 2024.

\bibitem[Ma et~al.(2024)Ma, Liu, Chen, Liu, Wu, Wu, Pan, Xie, Zhang, Zhao, et~al.]{janusflow}
Ma, Y., Liu, X., Chen, X., Liu, W., Wu, C., Wu, Z., Pan, Z., Xie, Z., Zhang, H., Zhao, L., et~al.
\newblock Janusflow: Harmonizing autoregression and rectified flow for unified multimodal understanding and generation.
\newblock \emph{arXiv preprint arXiv:2411.07975}, 2024.

\bibitem[McKinzie et~al.(2024)McKinzie, Gan, Fauconnier, Dodge, Zhang, Dufter, Shah, Du, Peng, Weers, et~al.]{mckinzie2024mm1}
McKinzie, B., Gan, Z., Fauconnier, J.-P., Dodge, S., Zhang, B., Dufter, P., Shah, D., Du, X., Peng, F., Weers, F., et~al.
\newblock Mm1: Methods, analysis \& insights from multimodal llm pre-training.
\newblock \emph{arXiv preprint arXiv:2403.09611}, 2024.

\bibitem[Murphy(2023)]{pml2Book}
Murphy, K.~P.
\newblock \emph{Probabilistic Machine Learning: Advanced Topics}.
\newblock MIT Press, 2023.
\newblock URL \url{http://probml.github.io/book2}.

\bibitem[Nichol et~al.(2021)Nichol, Dhariwal, Ramesh, Shyam, Mishkin, McGrew, Sutskever, and Chen]{nichol2021glide}
Nichol, A., Dhariwal, P., Ramesh, A., Shyam, P., Mishkin, P., McGrew, B., Sutskever, I., and Chen, M.
\newblock Glide: Towards photorealistic image generation and editing with text-guided diffusion models.
\newblock \emph{arXiv preprint arXiv:2112.10741}, 2021.

\bibitem[Parmar et~al.(2018)Parmar, Vaswani, Uszkoreit, Kaiser, Shazeer, Ku, and Tran]{parmar2018image}
Parmar, N., Vaswani, A., Uszkoreit, J., Kaiser, L., Shazeer, N., Ku, A., and Tran, D.
\newblock Image transformer.
\newblock In \emph{ICML}, pp.\  4055--4064, 2018.

\bibitem[Peebles \& Xie(2023{\natexlab{a}})Peebles and Xie]{dit}
Peebles, W. and Xie, S.
\newblock Scalable diffusion models with transformers.
\newblock In \emph{{ICCV}}, pp.\  4172--4182. {IEEE}, 2023{\natexlab{a}}.

\bibitem[Peebles \& Xie(2023{\natexlab{b}})Peebles and Xie]{peebles2023scalable}
Peebles, W. and Xie, S.
\newblock Scalable diffusion models with transformers.
\newblock In \emph{ICCV}, pp.\  4195--4205, 2023{\natexlab{b}}.

\bibitem[Penedo et~al.(2023)Penedo, Malartic, Hesslow, Cojocaru, Alobeidli, Cappelli, Pannier, Almazrouei, and Launay]{refinedweb}
Penedo, G., Malartic, Q., Hesslow, D., Cojocaru, R., Alobeidli, H., Cappelli, A., Pannier, B., Almazrouei, E., and Launay, J.
\newblock The refinedweb dataset for falcon {LLM:} outperforming curated corpora with web data only.
\newblock In \emph{NeurIPS}, 2023.

\bibitem[Peng et~al.(2023)Peng, Wang, Dong, Hao, Huang, Ma, and Wei]{Kosmos-2}
Peng, Z., Wang, W., Dong, L., Hao, Y., Huang, S., Ma, S., and Wei, F.
\newblock Kosmos-2: Grounding multimodal large language models to the world.
\newblock \emph{CoRR}, abs/2306.14824, 2023.

\bibitem[Podell et~al.(2023)Podell, English, Lacey, Blattmann, Dockhorn, M{\"u}ller, Penna, and Rombach]{sdxl}
Podell, D., English, Z., Lacey, K., Blattmann, A., Dockhorn, T., M{\"u}ller, J., Penna, J., and Rombach, R.
\newblock Sdxl: Improving latent diffusion models for high-resolution image synthesis.
\newblock \emph{arXiv preprint arXiv:2307.01952}, 2023.

\bibitem[Qu et~al.(2024)Qu, Zhang, Liu, Wang, Jiang, Gao, Ye, Du, Yuan, and Wu]{tokenflow}
Qu, L., Zhang, H., Liu, Y., Wang, X., Jiang, Y., Gao, Y., Ye, H., Du, D.~K., Yuan, Z., and Wu, X.
\newblock Tokenflow: Unified image tokenizer for multimodal understanding and generation.
\newblock \emph{CoRR}, abs/2412.03069, 2024.

\bibitem[Radford et~al.(2018)Radford, Narasimhan, Salimans, Sutskever, et~al.]{gpt1}
Radford, A., Narasimhan, K., Salimans, T., Sutskever, I., et~al.
\newblock Improving language understanding by generative pre-training.
\newblock 2018.

\bibitem[Radford et~al.(2021)Radford, Kim, Hallacy, Ramesh, Goh, Agarwal, Sastry, Askell, Mishkin, Clark, Krueger, and Sutskever]{clip}
Radford, A., Kim, J.~W., Hallacy, C., Ramesh, A., Goh, G., Agarwal, S., Sastry, G., Askell, A., Mishkin, P., Clark, J., Krueger, G., and Sutskever, I.
\newblock Learning transferable visual models from natural language supervision.
\newblock In \emph{{ICML}}, pp.\  8748--8763, 2021.

\bibitem[Raffel et~al.(2020)Raffel, Shazeer, Roberts, Lee, Narang, Matena, Zhou, Li, and Liu]{raffel2020exploring}
Raffel, C., Shazeer, N., Roberts, A., Lee, K., Narang, S., Matena, M., Zhou, Y., Li, W., and Liu, P.~J.
\newblock Exploring the limits of transfer learning with a unified text-to-text transformer.
\newblock \emph{Journal of machine learning research}, 21\penalty0 (140):\penalty0 1--67, 2020.

\bibitem[Ramesh et~al.(2021)Ramesh, Pavlov, Goh, Gray, Voss, Radford, Chen, and Sutskever]{dalle}
Ramesh, A., Pavlov, M., Goh, G., Gray, S., Voss, C., Radford, A., Chen, M., and Sutskever, I.
\newblock Zero-shot text-to-image generation.
\newblock In \emph{ICML}, pp.\  8821--8831. Pmlr, 2021.

\bibitem[Ramesh et~al.(2022{\natexlab{a}})Ramesh, Dhariwal, Nichol, Chu, and Chen]{dalle2}
Ramesh, A., Dhariwal, P., Nichol, A., Chu, C., and Chen, M.
\newblock Hierarchical text-conditional image generation with {CLIP} latents.
\newblock \emph{CoRR}, abs/2204.06125, 2022{\natexlab{a}}.

\bibitem[Ramesh et~al.(2022{\natexlab{b}})Ramesh, Dhariwal, Nichol, Chu, and Chen]{ramesh2022hierarchical}
Ramesh, A., Dhariwal, P., Nichol, A., Chu, C., and Chen, M.
\newblock Hierarchical text-conditional image generation with clip latents.
\newblock \emph{arXiv preprint arXiv:2204.06125}, 1\penalty0 (2):\penalty0 3, 2022{\natexlab{b}}.

\bibitem[Raposo et~al.(2024)Raposo, Ritter, Richards, Lillicrap, Humphreys, and Santoro]{mod}
Raposo, D., Ritter, S., Richards, B., Lillicrap, T., Humphreys, P.~C., and Santoro, A.
\newblock Mixture-of-depths: Dynamically allocating compute in transformer-based language models.
\newblock \emph{arXiv preprint arXiv:2404.02258}, 2024.

\bibitem[Ravuri \& Vinyals(2019)Ravuri and Vinyals]{ravuri2019classification}
Ravuri, S. and Vinyals, O.
\newblock Classification accuracy score for conditional generative models.
\newblock \emph{NeurIPS}, 32, 2019.

\bibitem[Rombach et~al.(2022)Rombach, Blattmann, Lorenz, Esser, and Ommer]{rombach2022high}
Rombach, R., Blattmann, A., Lorenz, D., Esser, P., and Ommer, B.
\newblock High-resolution image synthesis with latent diffusion models.
\newblock In \emph{CVPR}, pp.\  10684--10695, 2022.

\bibitem[Saharia et~al.(2022)Saharia, Chan, Saxena, Li, Whang, Denton, Ghasemipour, Gontijo~Lopes, Karagol~Ayan, Salimans, et~al.]{imagen}
Saharia, C., Chan, W., Saxena, S., Li, L., Whang, J., Denton, E.~L., Ghasemipour, K., Gontijo~Lopes, R., Karagol~Ayan, B., Salimans, T., et~al.
\newblock Photorealistic text-to-image diffusion models with deep language understanding.
\newblock \emph{NeurIPS}, 35:\penalty0 36479--36494, 2022.

\bibitem[Shen et~al.(2024)Shen, Song, Zhou, Chen, Li, Gong, Zhang, Tan, Kuen, Ding, et~al.]{lazydit}
Shen, X., Song, Z., Zhou, Y., Chen, B., Li, Y., Gong, Y., Zhang, K., Tan, H., Kuen, J., Ding, H., et~al.
\newblock Lazydit: Lazy learning for the acceleration of diffusion transformers.
\newblock \emph{arXiv preprint arXiv:2412.12444}, 2024.

\bibitem[Shi et~al.(2024)Shi, Han, Zhou, Liang, Lin, Zettlemoyer, and Yu]{llamafusion}
Shi, W., Han, X., Zhou, C., Liang, W., Lin, X.~V., Zettlemoyer, L., and Yu, L.
\newblock Llamafusion: Adapting pretrained language models for multimodal generation.
\newblock \emph{arXiv preprint arXiv:2412.15188}, 2024.

\bibitem[Sohl-Dickstein et~al.(2015)Sohl-Dickstein, Weiss, Maheswaranathan, and Ganguli]{sohl2015deep}
Sohl-Dickstein, J., Weiss, E., Maheswaranathan, N., and Ganguli, S.
\newblock Deep unsupervised learning using nonequilibrium thermodynamics.
\newblock In \emph{ICML}, pp.\  2256--2265, 2015.

\bibitem[Sou\v{c}ek et~al.(2024)Sou\v{c}ek, Damen, Wray, Laptev, and Sivic]{genhowto}
Sou\v{c}ek, T., Damen, D., Wray, M., Laptev, I., and Sivic, J.
\newblock Genhowto: Learning to generate actions and state transformations from instructional videos.
\newblock In \emph{CVPR}, pp.\  6561--6571, 2024.

\bibitem[Sun et~al.(2023{\natexlab{a}})Sun, Pan, Ge, Li, Duan, Wu, Zhang, Zhou, Qin, Wang, Dai, Qiao, Wang, and Li]{journeydb}
Sun, K., Pan, J., Ge, Y., Li, H., Duan, H., Wu, X., Zhang, R., Zhou, A., Qin, Z., Wang, Y., Dai, J., Qiao, Y., Wang, L., and Li, H.
\newblock Journeydb: {A} benchmark for generative image understanding.
\newblock In \emph{NeurIPS}, 2023{\natexlab{a}}.

\bibitem[Sun et~al.(2024)Sun, Jiang, Chen, Zhang, Peng, Luo, and Yuan]{llamagen}
Sun, P., Jiang, Y., Chen, S., Zhang, S., Peng, B., Luo, P., and Yuan, Z.
\newblock Autoregressive model beats diffusion: Llama for scalable image generation.
\newblock \emph{arXiv preprint arXiv:2406.06525}, 2024.

\bibitem[Sun et~al.(2023{\natexlab{b}})Sun, Cui, Zhang, Zhang, Yu, Luo, Wang, Rao, Liu, Huang, and Wang]{DBLP:journals/corr/abs-2312-13286}
Sun, Q., Cui, Y., Zhang, X., Zhang, F., Yu, Q., Luo, Z., Wang, Y., Rao, Y., Liu, J., Huang, T., and Wang, X.
\newblock Generative multimodal models are in-context learners.
\newblock \emph{CoRR}, abs/2312.13286, 2023{\natexlab{b}}.

\bibitem[Sun et~al.(2023{\natexlab{c}})Sun, Yu, Cui, Zhang, Zhang, Wang, Gao, Liu, Huang, and Wang]{DBLP:journals/corr/abs-2307-05222}
Sun, Q., Yu, Q., Cui, Y., Zhang, F., Zhang, X., Wang, Y., Gao, H., Liu, J., Huang, T., and Wang, X.
\newblock Generative pretraining in multimodality.
\newblock \emph{CoRR}, abs/2307.05222, 2023{\natexlab{c}}.

\bibitem[Sun et~al.(2023{\natexlab{d}})Sun, Yu, Cui, Zhang, Zhang, Wang, Gao, Liu, Huang, and Wang]{sun2023emu}
Sun, Q., Yu, Q., Cui, Y., Zhang, F., Zhang, X., Wang, Y., Gao, H., Liu, J., Huang, T., and Wang, X.
\newblock Emu: Generative pretraining in multimodality.
\newblock In \emph{ICLR}, 2023{\natexlab{d}}.

\bibitem[Tang et~al.(2024)Tang, Yang, Zhu, Zeng, and Bansal]{CoDI}
Tang, Z., Yang, Z., Zhu, C., Zeng, M., and Bansal, M.
\newblock Any-to-any generation via composable diffusion.
\newblock \emph{NeurIPS}, 36, 2024.

\bibitem[Team(2024)]{team2024chameleon}
Team, C.
\newblock Chameleon: Mixed-modal early-fusion foundation models.
\newblock \emph{arXiv preprint arXiv:2405.09818}, 2024.

\bibitem[Tong et~al.(2024{\natexlab{a}})Tong, Brown, Wu, Woo, Middepogu, Akula, Yang, Yang, Iyer, Pan, Wang, Fergus, LeCun, and Xie]{cambrian}
Tong, S., Brown, E., Wu, P., Woo, S., Middepogu, M., Akula, S.~C., Yang, J., Yang, S., Iyer, A., Pan, X., Wang, A., Fergus, R., LeCun, Y., and Xie, S.
\newblock Cambrian-1: {A} fully open, vision-centric exploration of multimodal llms.
\newblock \emph{CoRR}, abs/2406.16860, 2024{\natexlab{a}}.

\bibitem[Tong et~al.(2024{\natexlab{b}})Tong, Brown, Wu, Woo, Middepogu, Akula, Yang, Yang, Iyer, Pan, et~al.]{tong2024cambrian}
Tong, S., Brown, E., Wu, P., Woo, S., Middepogu, M., Akula, S.~C., Yang, J., Yang, S., Iyer, A., Pan, X., et~al.
\newblock Cambrian-1: A fully open, vision-centric exploration of multimodal llms.
\newblock \emph{arXiv preprint arXiv:2406.16860}, 2024{\natexlab{b}}.

\bibitem[Tong et~al.(2024{\natexlab{c}})Tong, Fan, Zhu, Xiong, Chen, Sinha, Rabbat, LeCun, Xie, and Liu]{metamorph}
Tong, S., Fan, D., Zhu, J., Xiong, Y., Chen, X., Sinha, K., Rabbat, M., LeCun, Y., Xie, S., and Liu, Z.
\newblock Metamorph: Multimodal understanding and generation via instruction tuning.
\newblock \emph{arXiv preprint arXiv:2412.14164}, 2024{\natexlab{c}}.

\bibitem[Touvron et~al.(2023)Touvron, Lavril, Izacard, Martinet, Lachaux, Lacroix, Rozi{\`{e}}re, Goyal, Hambro, Azhar, Rodriguez, Joulin, Grave, and Lample]{llama}
Touvron, H., Lavril, T., Izacard, G., Martinet, X., Lachaux, M., Lacroix, T., Rozi{\`{e}}re, B., Goyal, N., Hambro, E., Azhar, F., Rodriguez, A., Joulin, A., Grave, E., and Lample, G.
\newblock Llama: Open and efficient foundation language models.
\newblock \emph{CoRR}, abs/2302.13971, 2023.

\bibitem[Van Den~Oord et~al.(2017)Van Den~Oord, Vinyals, et~al.]{vqvae}
Van Den~Oord, A., Vinyals, O., et~al.
\newblock Neural discrete representation learning.
\newblock \emph{NeurIPS}, 30, 2017.

\bibitem[Vaswani et~al.(2017{\natexlab{a}})Vaswani, Shazeer, Parmar, Uszkoreit, Jones, Gomez, Kaiser, and Polosukhin]{attention}
Vaswani, A., Shazeer, N., Parmar, N., Uszkoreit, J., Jones, L., Gomez, A.~N., Kaiser, {\L}., and Polosukhin, I.
\newblock Attention is all you need.
\newblock \emph{NeurIPS}, 30, 2017{\natexlab{a}}.

\bibitem[Vaswani et~al.(2017{\natexlab{b}})Vaswani, Shazeer, Parmar, Uszkoreit, Jones, Gomez, Kaiser, and Polosukhin]{transformer}
Vaswani, A., Shazeer, N., Parmar, N., Uszkoreit, J., Jones, L., Gomez, A.~N., Kaiser, {\L}., and Polosukhin, I.
\newblock Attention is all you need.
\newblock \emph{NeurIPS}, 2017{\natexlab{b}}.

\bibitem[Wang \& Cho(2019)Wang and Cho]{wang2019bert}
Wang, A. and Cho, K.
\newblock Bert has a mouth, and it must speak: Bert as a markov random field language model.
\newblock \emph{arXiv preprint arXiv:1902.04094}, 2019.

\bibitem[Wang et~al.(2024)Wang, Zhang, Luo, Sun, Cui, Wang, Zhang, Wang, Li, Yu, et~al.]{emu3}
Wang, X., Zhang, X., Luo, Z., Sun, Q., Cui, Y., Wang, J., Zhang, F., Wang, Y., Li, Z., Yu, Q., et~al.
\newblock Emu3: Next-token prediction is all you need.
\newblock \emph{arXiv preprint arXiv:2409.18869}, 2024.

\bibitem[Wortsman et~al.(2023)Wortsman, Liu, Xiao, Everett, Alemi, Adlam, Co-Reyes, Gur, Kumar, Novak, et~al.]{qknorm2}
Wortsman, M., Liu, P.~J., Xiao, L., Everett, K., Alemi, A., Adlam, B., Co-Reyes, J.~D., Gur, I., Kumar, A., Novak, R., et~al.
\newblock Small-scale proxies for large-scale transformer training instabilities.
\newblock \emph{arXiv preprint arXiv:2309.14322}, 2023.

\bibitem[Wu et~al.(2024{\natexlab{a}})Wu, Jiang, Ma, Liu, Zhao, Yuan, Bai, and Bai]{liquid}
Wu, J., Jiang, Y., Ma, C., Liu, Y., Zhao, H., Yuan, Z., Bai, S., and Bai, X.
\newblock Liquid: Language models are scalable multi-modal generators.
\newblock \emph{CoRR}, abs/2412.04332, 2024{\natexlab{a}}.

\bibitem[Wu et~al.(2023{\natexlab{a}})Wu, Ge, Wang, Lei, Gu, Shi, Hsu, Shan, Qie, and Shou]{wu2023tune}
Wu, J.~Z., Ge, Y., Wang, X., Lei, S.~W., Gu, Y., Shi, Y., Hsu, W., Shan, Y., Qie, X., and Shou, M.~Z.
\newblock Tune-a-video: One-shot tuning of image diffusion models for text-to-video generation.
\newblock In \emph{ICCV}, 2023{\natexlab{a}}.

\bibitem[Wu et~al.(2023{\natexlab{b}})Wu, Fei, Qu, Ji, and Chua]{wu2023next}
Wu, S., Fei, H., Qu, L., Ji, W., and Chua, T.-S.
\newblock Next-gpt: Any-to-any multimodal llm.
\newblock \emph{arXiv preprint arXiv:2309.05519}, 2023{\natexlab{b}}.

\bibitem[Wu et~al.(2024{\natexlab{b}})Wu, Chen, Lin, Wang, Gao, Xu, Xu, Hu, Chen, and Shou]{videollmmod}
Wu, S., Chen, J., Lin, K.~Q., Wang, Q., Gao, Y., Xu, Q., Xu, T., Hu, Y., Chen, E., and Shou, M.~Z.
\newblock Videollm-mod: Efficient video-language streaming with mixture-of-depths vision computation.
\newblock \emph{CoRR}, abs/2408.16730, 2024{\natexlab{b}}.

\bibitem[Wu et~al.(2024{\natexlab{c}})Wu, Zhang, Chen, Tang, Li, Fang, Zhu, Xie, Yin, Yi, Han, and Lu]{vila-u}
Wu, Y., Zhang, Z., Chen, J., Tang, H., Li, D., Fang, Y., Zhu, L., Xie, E., Yin, H., Yi, L., Han, S., and Lu, Y.
\newblock {VILA-U:} a unified foundation model integrating visual understanding and generation.
\newblock \emph{CoRR}, abs/2409.04429, 2024{\natexlab{c}}.

\bibitem[Xiao et~al.(2024)Xiao, Tian, Chen, Han, and Lewis]{attentionsink}
Xiao, G., Tian, Y., Chen, B., Han, S., and Lewis, M.
\newblock Efficient streaming language models with attention sinks.
\newblock In \emph{{ICLR}}. OpenReview.net, 2024.

\bibitem[Xie et~al.(2024)Xie, Mao, Bai, Zhang, Wang, Lin, Gu, Chen, Yang, and Shou]{showo}
Xie, J., Mao, W., Bai, Z., Zhang, D.~J., Wang, W., Lin, K.~Q., Gu, Y., Chen, Z., Yang, Z., and Shou, M.~Z.
\newblock Show-o: One single transformer to unify multimodal understanding and generation.
\newblock \emph{arXiv preprint arXiv:2408.12528}, 2024.

\bibitem[Xue et~al.(2024{\natexlab{a}})Xue, Zheng, Fu, Ni, Zheng, Zhou, and You]{openmoe}
Xue, F., Zheng, Z., Fu, Y., Ni, J., Zheng, Z., Zhou, W., and You, Y.
\newblock Openmoe: An early effort on open mixture-of-experts language models.
\newblock In \emph{{ICML}}. OpenReview.net, 2024{\natexlab{a}}.

\bibitem[Xue et~al.(2024{\natexlab{b}})Xue, Song, Guo, Liu, Zong, Liu, and Luo]{raphael}
Xue, Z., Song, G., Guo, Q., Liu, B., Zong, Z., Liu, Y., and Luo, P.
\newblock Raphael: Text-to-image generation via large mixture of diffusion paths.
\newblock \emph{NeurIPS}, 36, 2024{\natexlab{b}}.

\bibitem[Yang et~al.(2024)Yang, Ge, Li, Chen, Ge, Shan, and Chen]{seed-story}
Yang, S., Ge, Y., Li, Y., Chen, Y., Ge, Y., Shan, Y., and Chen, Y.
\newblock Seed-story: Multimodal long story generation with large language model.
\newblock \emph{arXiv preprint arXiv:2407.08683}, 2024.

\bibitem[Ye et~al.(2024{\natexlab{a}})Ye, Huang, Lu, Yu, Ping, Tao, Kautz, Han, Xu, Molchanov, et~al.]{x-vila}
Ye, H., Huang, D.-A., Lu, Y., Yu, Z., Ping, W., Tao, A., Kautz, J., Han, S., Xu, D., Molchanov, P., et~al.
\newblock X-vila: Cross-modality alignment for large language model.
\newblock \emph{arXiv preprint arXiv:2405.19335}, 2024{\natexlab{a}}.

\bibitem[Ye et~al.(2024{\natexlab{b}})Ye, Xu, Ye, Yan, Hu, Liu, Qian, Zhang, and Huang]{ye2024mplug}
Ye, Q., Xu, H., Ye, J., Yan, M., Hu, A., Liu, H., Qian, Q., Zhang, J., and Huang, F.
\newblock mplug-owl2: Revolutionizing multi-modal large language model with modality collaboration.
\newblock In \emph{CVPR}, pp.\  13040--13051, 2024{\natexlab{b}}.

\bibitem[You et~al.(2024)You, Barnes, Zhou, Kang, Du, Zhou, Zhang, Nitzan, Liu, Lin, et~al.]{layermoddit}
You, H., Barnes, C., Zhou, Y., Kang, Y., Du, Z., Zhou, W., Zhang, L., Nitzan, Y., Liu, X., Lin, Z., et~al.
\newblock Layer-and timestep-adaptive differentiable token compression ratios for efficient diffusion transformers.
\newblock \emph{arXiv preprint arXiv:2412.16822}, 2024.

\bibitem[Young et~al.(2014)Young, Lai, Hodosh, and Hockenmaier]{flickre}
Young, P., Lai, A., Hodosh, M., and Hockenmaier, J.
\newblock From image descriptions to visual denotations: New similarity metrics for semantic inference over event descriptions.
\newblock \emph{Trans. Assoc. Comput. Linguistics}, 2:\penalty0 67--78, 2014.

\bibitem[Yu et~al.(2021)Yu, Li, Koh, Zhang, Pang, Qin, Ku, Xu, Baldridge, and Wu]{parti}
Yu, J., Li, X., Koh, J.~Y., Zhang, H., Pang, R., Qin, J., Ku, A., Xu, Y., Baldridge, J., and Wu, Y.
\newblock Vector-quantized image modeling with improved vqgan.
\newblock \emph{arXiv preprint arXiv:2110.04627}, 2021.

\bibitem[Yu et~al.(2022)Yu, Xu, Koh, Luong, Baid, Wang, Vasudevan, Ku, Yang, Ayan, et~al.]{yu2022scaling}
Yu, J., Xu, Y., Koh, J.~Y., Luong, T., Baid, G., Wang, Z., Vasudevan, V., Ku, A., Yang, Y., Ayan, B.~K., et~al.
\newblock Scaling autoregressive models for content-rich text-to-image generation.
\newblock \emph{arXiv preprint arXiv:2206.10789}, 2\penalty0 (3):\penalty0 5, 2022.

\bibitem[Yu et~al.(2023)Yu, Lezama, Gundavarapu, Versari, Sohn, Minnen, Cheng, Gupta, Gu, Hauptmann, et~al.]{magvitv2}
Yu, L., Lezama, J., Gundavarapu, N.~B., Versari, L., Sohn, K., Minnen, D., Cheng, Y., Gupta, A., Gu, X., Hauptmann, A.~G., et~al.
\newblock Language model beats diffusion--tokenizer is key to visual generation.
\newblock \emph{arXiv preprint arXiv:2310.05737}, 2023.

\bibitem[Yue et~al.(2024)Yue, Ni, Zheng, Zhang, Liu, Zhang, Stevens, Jiang, Ren, Sun, Wei, Yu, Yuan, Sun, Yin, Zheng, Yang, Liu, Huang, Sun, Su, and Chen]{mmmu}
Yue, X., Ni, Y., Zheng, T., Zhang, K., Liu, R., Zhang, G., Stevens, S., Jiang, D., Ren, W., Sun, Y., Wei, C., Yu, B., Yuan, R., Sun, R., Yin, M., Zheng, B., Yang, Z., Liu, Y., Huang, W., Sun, H., Su, Y., and Chen, W.
\newblock {MMMU:} {A} massive multi-discipline multimodal understanding and reasoning benchmark for expert {AGI}.
\newblock In \emph{{CVPR}}, pp.\  9556--9567. {IEEE}, 2024.

\bibitem[Zeng et~al.(2023)Zeng, Du, Wang, Xu, Lei, Chen, and Cui]{skip}
Zeng, D., Du, N., Wang, T., Xu, Y., Lei, T., Chen, Z., and Cui, C.
\newblock Learning to skip for language modeling.
\newblock \emph{CoRR}, abs/2311.15436, 2023.

\bibitem[Zhang et~al.(2024{\natexlab{a}})Zhang, Meng, Qi, Huang, Wu, and Wang]{pmod}
Zhang, J., Meng, D., Qi, J., Huang, Z., Wu, T., and Wang, L.
\newblock p-mod: Building mixture-of-depths mllms via progressive ratio decay.
\newblock \emph{CoRR}, abs/2412.04449, 2024{\natexlab{a}}.

\bibitem[Zhang et~al.(2024{\natexlab{b}})Zhang, Xiong, Yang, Casas, Hu, and Urtasun]{copilot4d}
Zhang, L., Xiong, Y., Yang, Z., Casas, S., Hu, R., and Urtasun, R.
\newblock Copilot4d: Learning unsupervised world models for autonomous driving via discrete diffusion.
\newblock In \emph{ICLR}, 2024{\natexlab{b}}.

\bibitem[Zhou et~al.(2024)Zhou, Yu, Babu, Tirumala, Yasunaga, Shamis, Kahn, Ma, Zettlemoyer, and Levy]{transfusion}
Zhou, C., Yu, L., Babu, A., Tirumala, K., Yasunaga, M., Shamis, L., Kahn, J., Ma, X., Zettlemoyer, L., and Levy, O.
\newblock Transfusion: Predict the next token and diffuse images with one multi-modal model.
\newblock \emph{CoRR}, abs/2408.11039, 2024.

\bibitem[Zhu et~al.(2023{\natexlab{a}})Zhu, Chen, Shen, Li, and Elhoseiny]{minigpt4}
Zhu, D., Chen, J., Shen, X., Li, X., and Elhoseiny, M.
\newblock Minigpt-4: Enhancing vision-language understanding with advanced large language models.
\newblock \emph{CoRR}, abs/2304.10592, 2023{\natexlab{a}}.

\bibitem[Zhu et~al.(2023{\natexlab{b}})Zhu, Ding, Ge, Ge, Zhao, Zhao, Wang, and Shan]{vlgpt}
Zhu, J., Ding, X., Ge, Y., Ge, Y., Zhao, S., Zhao, H., Wang, X., and Shan, Y.
\newblock {VL-GPT:} {A} generative pre-trained transformer for vision and language understanding and generation.
\newblock \emph{CoRR}, abs/2312.09251, 2023{\natexlab{b}}.

\end{thebibliography}
\bibliographystyle{icml2025}


\newpage
\appendix
\onecolumn
\section{Appendix}


In this appendix, we provide additional implementation details of our experiments (Sec.~\ref{showo_dataset}), discussion about the training cost(Sec.~\ref{discussion_costs}), the results of adapting MoD to diffusion models (Sec.~\ref{adaption_to_diffusion}), and the formula for the Straight-Through Gumbel Softmax (Sec.~\ref{gumbel_softmax_formula}).

\subsection{More Implementation Details} 
\label{showo_dataset}
\textbf{Dataset Enhancement and Streamlined Training Workflow in the Show-o Model.} The original Show-o model consists of multiple training stages. In the first two stages, it is trained on the ImageNet~\cite{imagenet} dataset and large-scale text-image paired data to achieve effective text-image alignment. The third stage leverages high-quality data to develop generation capabilities, while the final two stages utilize the LLaVA dataset~\cite{llava} to enhance understanding capabilities.

In this work, we improve the Show-o training pipeline by incorporating additional understanding datasets and reducing the training process to two stages. The original Show-o exclusively used the LLaVA dataset~\cite{llava} for training its understanding component. However, data imbalance caused the model to develop generation capabilities before understanding capabilities, which adversely affected generation quality. To resolve this, we introduce the Cambrian dataset~\cite{cambrian} and internal high-quality data to fine-tune the Show-o model within a two-stage training framework. The model processes images at a resolution of 512×512 pixels. Our revised pipeline achieves comparable results while reducing computational resource usage. We reevaluate the primary benchmarks and compare them with the original Show-o results.

\textbf{Emu3.} The Emu3 public repository does not include the training code or data for MMU tasks. To address this, we utilize the LLaVA-v1.5-mix-665K dataset to represent MMU capabilities and enhance the training pipeline with additional MMU-specific code. We employ the same generation data as the Show-o model and fine-tune Emu3 for 2 epochs with a learning rate of 2e-5, processing images at a resolution of 512×512 pixels.

In our experiments, we use the Show-o and Emu3 checkpoints from the first training stage, which involves low-quality text-to-image generation and limited captioning capabilities, as base models. To validate our method, we fine-tune these models with high-quality image data and understanding QA data. The router utilizes a single linear network.

\subsection{Discussion on Training Costs}
\label{discussion_costs}

As shown in Tab.~\ref{tab:training_costs}, although Emu3 prunes more tokens and achieves greater reductions in FLOPs and training speed, its memory usage remains largely unchanged compared to the full computation model. This discrepancy arises from the significant size difference between Emu3 and Show-o, with Emu3 having 8.5B parameters compared to Show-o's 1.4B. In Emu3, the large number of parameters and their corresponding gradients dominate memory consumption, making token pruning's impact on memory minimal. Instead, pruning primarily enhances training speed because the extensive parameter count leads to substantial attention computation per layer, making token reduction more impactful on speed. In contrast, the smaller Show-o model allocates a substantial portion of memory to activation variables during token computation. Consequently, token pruning in Show-o leads to a more significant reduction in memory usage.

\begin{table*}[h]
\centering
\resizebox{\linewidth}{!}{ 
\begin{tabular}{llccccccc|c}
        \toprule
    	Model & Method & TFLOPS$\downarrow$ $\downarrow$ & Single Obj. & Two Obj. & Counting  & Colors &  Position & Color Attri. & Overall$\uparrow$ \\
        \midrule
            \multirow{3}{*}{PixArt-alpha}
            & Original Model & 42.64 &0.98 &0.50 &0.44 &0.80 &0.08 &0.07 &0.48\\
            & Full Computation & 42.64  &0.98 & 0.71 & 0.55 & 0.82 & 0.17 & 0.29&  0.58 \\
            & Interleaved Layer &32.03   &0.98 & 0.59 & 0.43 & 0.80 & 0.13 & 0.22& 0.53 \\
             & UniMoD &  32.03  &0.99 & 0.68 & 0.47 & 0.81 & 0.17 & 0.26 & 0.56 \\
        \bottomrule

\end{tabular}
}

\caption{\method{} for PixArt. \method{} achieves results closest to the full computation model.}

\label{tab:table_pixart}
\end{table*}

\subsection{Adaptation to Diffusion Models} 
\label{adaption_to_diffusion}
\textbf{MoD for Generation Models.} While our method is primarily designed for unified transformers, we also validate its effectiveness for training or fine-tuning generation models such as DiT~\cite{dit} and PixArt~\cite{pixart}. For the DiT model, we select the checkpoint trained for 50k iterations as the base model to calculate the ARank. For the PixArt model, we use the final model as the base and fine-tune it using our internal high-quality data (used in the Show-o fine-tuning stage). We evaluate the DiT model using the ImageNet~\cite{imagenet} FID metric, and for PixArt, we use the GenEval benchmark~\cite{geneval} to assess generation quality. Both models process images at a resolution of 256×256 pixels. 

For PixArt, we conduct three experiments. First, we present the results using full computation. Second, we prune 40\% of the tokens in the interleaved layers. Finally, we calculate the ARank value for each layer and select the 14 layers with the lowest ARank values to prune 40\% of the tokens. As shown in Tab.~\ref{tab:table_pixart}, our ARank-based MoD method achieves better performance compared to standard pruning approaches at the same computational cost, with only a slight reduction in performance relative to full computation.

\begin{wraptable}{r}{0.5\textwidth}
\centering
\resizebox{\linewidth}{!}{ 
\begin{tabular}{llcccccc}
        \toprule
    	Model & Method & TFLOPS$\downarrow$& IS$\uparrow$ & FID$\downarrow$ \\
         \midrule
            \multirow{2}{*}{DiT-XL/2}
            & Full Computation & 117.2  & 168.16 &4.97\\
             & UniMoD & 93.6  & 171.67 &5.45\\
        \bottomrule

\end{tabular}
}

\caption{UniMoD for DiT. UniMoD achieves similar results as the original model after 500K iterations.}
\label{tab:table_dit}
\end{wraptable}

For DiT, we select the checkpoint trained for 50k iterations as the base model to calculate the ARank. Then we choose the least value  14 layers to prune tokens. In practice, we gradually scale the token capacity from 1 to 0.2 over the course of 500k training iterations. Finally, we evaluate the model using a capacity of 0.4, as shown in Tab.~\ref{tab:table_dit}.

From these experiments, we observe that the token sequences in both DiT and PixArt exhibit higher redundancy compared to those in Show-o. We attribute this difference to the design of their image tokenizers. The VAEs used in DiT and PixArt downsample images by a factor of 8, whereas Show-o's tokenizer downsamples by a factor of 16. Consequently, for images of the same resolution, DiT and PixArt require more tokens to represent the image than Show-o.

\subsection{Straight-Through Gumbel Softmax}
\label{gumbel_softmax_formula}

The Straight-Through Gumbel-Softmax method assigns binary weights to tokens, allowing discrete sampling while preserving differentiability through a straight-through estimator essential for gradient-based optimization. In the forward pass, tokens with the highest probability are selected for retention or pruning. During the backward pass, soft probabilities are used to compute gradients, allowing smooth updates.

The formula for the Gumbel Softmax is:
\begin{equation}
y_i = \frac{\exp\left( \dfrac{\log(\pi_i) + g_i}{\tau} \right)}{\sum\limits_{j=1}^{K} \exp\left( \dfrac{\log(\pi_j) + g_j}{\tau} \right)}.
\end{equation}
$\pi_i$ is the original probability of the $i$-th category, $g_i$ is noise sampled from the $\text{Gumbel}(0,1)$ distribution, $\tau$ is the temperature parameter, and $K$ is the number of categories.

In the straight-through version, we obtain a hard one-hot vector z during the forward pass by taking the index with the highest y:
\begin{equation}
z_i = \begin{cases}
1, & \text{if } i = \arg\max\limits_j \, y_j \\
0, & \text{otherwise}
\end{cases}
\end{equation}

In this work, we employ this method to assess the importance of tokens across different tasks. The auxiliary loss is defined as \( \mathcal{L}_{\text{aux}} = P \sum_{i}(r_i - P)^2 \), where \( r_i \) is the capacity of the \( i \)-th layer (\( i \leq L \)). This loss function is utilized to ensure that approximately half of the tokens are processed by each layer.




\end{document}